\pdfoutput=1

\documentclass[11pt]{article}

\usepackage[final]{acl}

\usepackage{times}
\usepackage{latexsym}
\usepackage{times,latexsym}
\usepackage{url}
\usepackage[T1]{fontenc}
\usepackage{graphicx}
\usepackage{stmaryrd}
\usepackage{rotating}
\usepackage{multirow}
\usepackage{booktabs}
\usepackage{lscape}
\usepackage{hyperref}
\usepackage{tcolorbox}
\usepackage{adjustbox}
\usepackage{algorithm}
\usepackage{algorithmicx}
\usepackage{algpseudocode}
\usepackage{tablefootnote}
\usepackage{siunitx}
\usepackage{hyperref}
\usepackage{mathrsfs}
\usepackage{mathtools}
\usepackage{longtable}
\usepackage{amssymb}
\usepackage{verbatim}
\usepackage{amsmath}
\usepackage{multirow}
\usepackage{graphicx}
\usepackage{subcaption}
\usepackage{enumitem}
\usepackage[T1]{fontenc}

\usepackage[utf8]{inputenc}

\usepackage{microtype}

\usepackage{inconsolata}

\usepackage{graphicx}

\title{DRES: Fake news detection by dynamic representation and ensemble selection}

\author{
  \textbf{Faramarz Farhangian\textsuperscript{1}},
  \textbf{Leandro A. Ensina\textsuperscript{2}},
  \textbf{George D. C. Cavalcanti\textsuperscript{3}},
  \textbf{Rafael M. O. Cruz\textsuperscript{1}}
\\
\\
  \textsuperscript{1}École de Technologie Supérieure (ÉTS-Montréal), Canada \\
  \textsuperscript{2}Universidade Tecnológica Federal do Paraná (UTFPR), Brazil \\
  \textsuperscript{3}Universidade Federal de Pernambuco (UFPE), Brazil
\\
  \small{
    \href{mailto:faramarz.farhangian.1@ens.etsmtl.ca}{faramarz.farhangian.1@ens.etsmtl.ca}, 
    \href{mailto:leandroa@utfpr.edu.br}{leandroa@utfpr.edu.br}, 
    \href{mailto:gdcc@cin.ufpe.br}{gdcc@cin.ufpe.br}, 
    \href{mailto:rafael.menelau-cruz@etsmtl.ca}{rafael.menelau-cruz@etsmtl.ca}
  }
}

\begin{document}
\maketitle

\begin{abstract}
The rapid spread of information via social media has made text-based fake news detection critically important due to its societal impact. This paper presents a novel detection method called Dynamic Representation and Ensemble Selection (DRES) for identifying fake news based solely on text. DRES leverages instance hardness measures to estimate the classification difficulty for each news article across multiple textual feature representations. By dynamically selecting the textual representation and the most competent ensemble of classifiers for each instance, DRES significantly enhances prediction accuracy. Extensive experiments show that DRES achieves notable improvements over state-of-the-art methods, confirming the effectiveness of representation selection based on instance hardness and dynamic ensemble selection in boosting performance. Codes and
data are available at: \url{https://github.com/FFarhangian/FakeNewsDetection_DRES}.
\end{abstract}

\section{Introduction}

Detecting fake news is an increasingly important task in today’s world as false news spreads significantly faster and deeper than true news~\cite{vosoughi2018spread}. This problem is further exacerbated by 
generative AI, which amplifies misinformation by creating highly persuasive but fabricated content~\cite{loth2024blessing}. Traditional text-based models frequently struggle with context sensitivity and generalization, especially when processing ambiguous text or domain-shifted inputs~\cite{wang2017liar,reddy2020text}. Addressing these issues is essential to support the credibility of information in online platforms and public communication.

\begin{figure}[htp]
    \centering
    \includegraphics[width=0.45\textwidth]{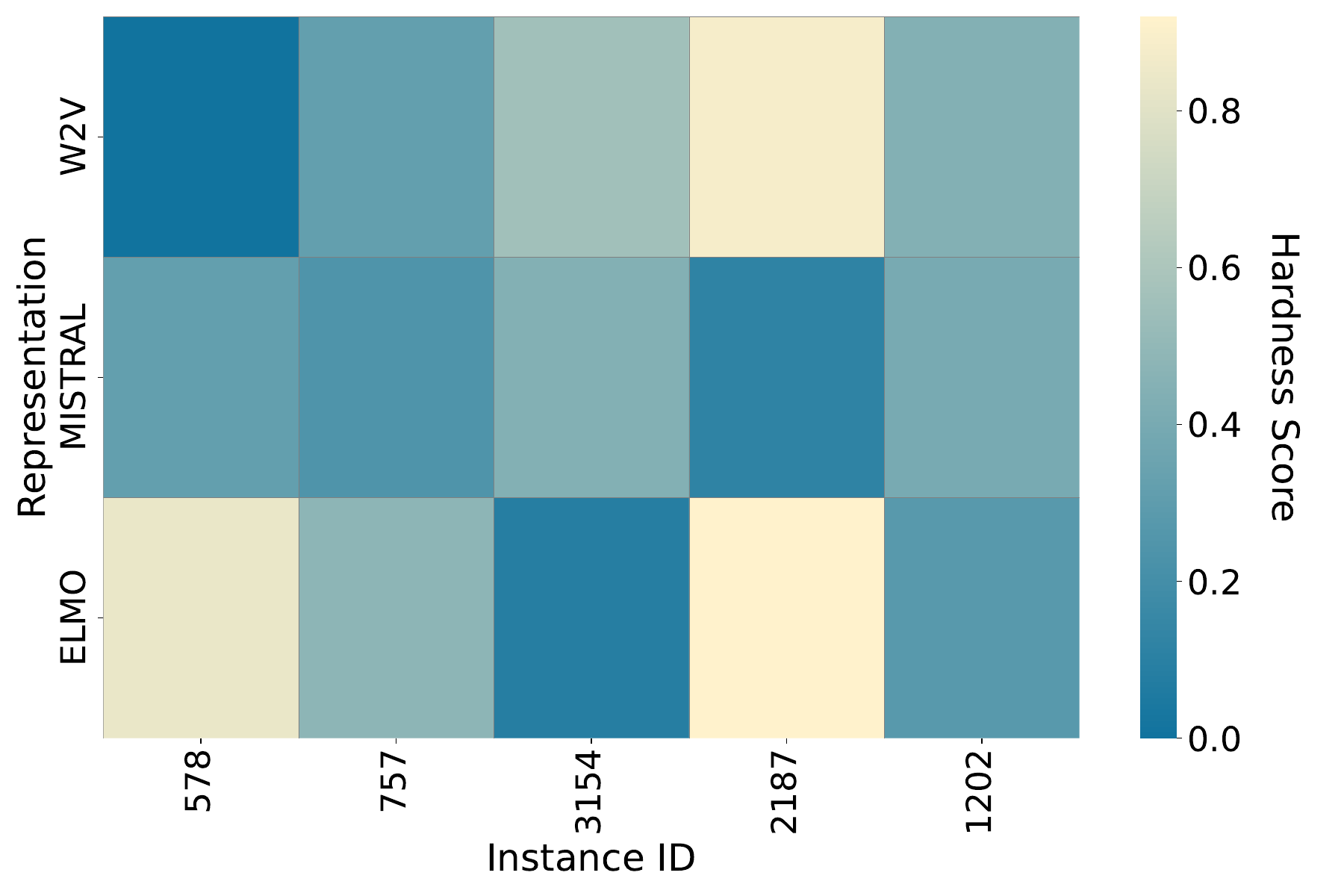}
    \caption{Instance hardness heat map for a few instances taken from the Liar dataset.}
    \label{fig:hardness_motivation}
    \vspace{-1em}
\end{figure}

While large language models (LLMs) such as Mistral~\cite{jiang2023mistral7b} have improved performance in many text classification tasks, relying on a single feature representation may be insufficient for fake news detection. Different representations capture complementary aspects of the input, such as surface-level statistics or contextual semantics. Recent work~\cite{farhangian2024fake} shows that using multiple representations, even with the same classifier, can reduce misclassification errors by exploiting this diversity. However, attempts to naively combine all representations (e.g., ~\cite{PCC2025}) often yield suboptimal results due to increased noise failing to account for input-specific representation effectiveness.

\begin{figure}[htp]
    \centering
    \includegraphics[width=0.495\textwidth]{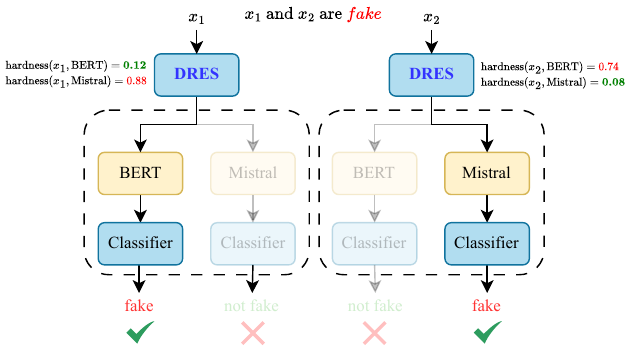}
    \caption{Overview of DRES that dynamically selects the most suitable text representation based on the test-time instance hardness.}
    \label{fig:motivations}
\end{figure}

In this paper, we argue that robust detection requires dynamic selection of both representations and classifiers based on input characteristics. We formalize this via instance hardness~\cite{smith2014instance}, which measures how likely a sample is to be misclassified by any learning algorithm under a given representation. Figure~\ref{fig:hardness_motivation} illustrates this, showing hardness scores for Liar dataset~\cite{wang2017liar} instances across three representations. A news item may have low hardness in ELMO's space but high hardness in Mistral's, with the reverse holding for other inputs. Thus, naively combining all representations or classifiers may hurt performance.

To bridge this gap, we propose a two-stage framework called \textbf{DRES} (Dynamic Representation and Ensemble Selection). An overview is shown in Figure~\ref{fig:motivations}. First, given an input, DRES estimates instance hardness across multiple text representations using a test-time variant of the k-Disagreeing Neighbors (kDN) metric~\cite{smith2014instance}. Based on this estimate, the representation with the lowest hardness is selected.

However, even within the chosen representation, classifiers trained on that space can still exhibit distinct local error patterns, especially for harder instances. To address this, DRES applies dynamic ensemble selection~\cite{cruz2018dynamic} as a second stage, selecting only the most competent classifiers in the neighborhood of the query rather than averaging over the entire pool. This ensures that predictions are adapted to the local region of the input, improving robustness on difficult cases.
The final prediction is obtained by majority voting over this selected subset. While prior work~\cite{cook2025no} used hardness metrics to guide training or data sampling, we extend their use to test time, showing that instance hardness can also support inference-time adaptation.

Our approach introduces a novel synergy between representation selection and classifier specialization. By first selecting the representation with the lowest estimated instance hardness, the framework maps the problem to a space where classifiers are more likely to agree. A second stage then refines this by dynamically selecting the most locally competent classifiers in that space per test instance. To the best of our knowledge, this is the first work to apply dynamic representation and ensemble selection jointly, and to repurpose complexity-based metrics to guide inference-time decisions.

The contributions of this paper are as follows: 1)~We propose a dynamic multi-view ensemble framework for fake news detection that adaptively selects both the most effective text representation and the most competent classifiers for each instance. 2)~We introduce a novel test-time estimation approach for instance hardness using a supervised metric (kDN) to guide representation selection decisions during inference. 3)~We empirically demonstrate that combining diverse text representations with dynamic classifier selection leads to consistent performance improvements across multiple fake news datasets while having plenty of potential for future improvements.

\section{Related work}

Fake news detection has been approached from diverse perspectives, including image-based analysis~\cite{qi2019exploiting} and social-context methods (i.e., modeling network structure and propagation patterns) with geometric deep learning~\cite{monti2019fake}. As this work is dedicated to textual content, our review focuses on text-only models and hybrid text-augmented methods.

\noindent \textbf{Text-Only Fake News Detection.} The earliest text-only methods relied on shallow lexical and stylometric features such as TF-IDF representations~\cite{patwa2021fighting}, n-grams, punctuation patterns, and readability metrics \cite{agudelo2018raising} fed into classifiers like Logistic Regression, SVM, or Random Forest \cite{shu2017fake}. \citet{perezrosas2018} expanded on these by adding psycholinguistic and rhetorical cues to capture deceptive language, while \citet{potthast2018stylometric} showed that pure stylometric features alone can separate hyperpartisan and fake news from genuine articles.

Later work moved to dense representations. Word-embedding–based models include BiLSTM over GloVe \cite{SASTRAWAN2022396}, and CNNs on Word2Vec \cite{girgis2018deep}. These were followed by transformer models like FakeBERT, which pairs BERT embeddings with 1D-CNN filters \cite{kaliyar2021fakebert}, ScrutNet’s Bi-LSTM + CNN fusion \cite{verma2025scrutnet}, and by attention-driven architectures such as 3HAN, a three-level hierarchical network that attends separately to words, sentences, and headlines for interpretable classification \cite{singhania20173han}. Finally, end-to-end transformer classifiers (e.g., BERT, LLaMA) now serve as strong baselines \cite{10074216,farhangian2024fake}. These unimodal approaches all use fixed representations or model combinations and cannot adjust to per-instance variations in text complexity, making them vulnerable to domain shifts and adversarial inputs.

\noindent \textbf{Multiple Text Representation Techniques.} Some approaches combine different textual representations of the same input to enrich the feature space. \citet{essa2023fake} merge several layers of BERT embeddings and use the result in a LightGBM classifier, while \citet{gautam2021covidxlnetlda} integrate XLNet embeddings with LDA topics to capture both contextual and thematic cues. MisRoBÆRTa \cite{truica2022} fuses BART and RoBERTa encodings into an ensemble architecture. MVAE \cite{PCC2025} aligns heterogeneous features into a shared latent space with Multi-View Auto Encoder. 

Nevertheless, these methods fuse all representations uniformly, without accounting for input-specific suitability. This can lead to inefficiencies and misclassification, as shown in \cite{PCC2025}, where joint representations from MVAE reduce performance on certain instances. Moreover, \citet{peng2024not} further highlight that news samples vary widely in structure, supporting the need for context-aware embedding selection. In contrast, the proposed DRES works by first estimating the hardness of each instance and dynamically selecting the most suitable representation before applying ensemble selection within that space. Thus, the representation and classifiers are adapted to the specific characteristics of each input.

\noindent \textbf{Hybrid text-augmented methods.} To harness richer signals, some methods integrate metadata (e.g., publication source, timestamps, user profiles) or fuse multiple data modalities. Social‑context models encode user engagement and propagation patterns via graph neural networks \cite{shu2019beyond}.
Multimodal frameworks process text alongside images using attention or contrastive learning—SpotFake aligns text and image embeddings~\cite{singhal2019spotfake}, while MCOT applies optimal transport for cross‑modal fusion \cite{zhou2024mcot}. Hybrid systems further incorporate temporal or social features into neural classifiers \cite{ruchansky2017csi} to improve classification. 

Multimodal approaches that integrate textual and social context features show consistent improvements in fake news detection accuracy~\cite{huang2019multimodal}, with late fusion methods combining text, image, and social signals providing further gains \cite{nguyen2019fang}. DANES~\cite{truicua2024danes} combines RNN-based text encoding with RNN-CNN social context encoding, concatenating both into a unified embedding for final classification via a dense layer. Metadata and multimodal inputs are often dataset‑specific and may be missing or noisy in real‑world settings, risking biased predictions and limiting generalization beyond curated benchmarks.

\noindent \textbf{Dynamic Ensemble Selection.} Ensemble methods improve performance by combining multiple base learners to reduce bias and variance compared to any single model \cite{dietterich2000ensemble}. Static ensembles use fixed strategies such as majority voting \cite{kittler1998combining} or stacked generalization \cite{wolpert1992stacked}. Dynamic ensemble selection (DES) instead adapts to each query by (1) defining a Region of Competence (RoC) around the sample (e.g., via clustering or similarity search), (2) measuring each classifier’s competence within that RoC, and (3) combining the outputs of the most competent models. DES techniques mainly differ in the heuristics used to estimate classifier competence within the Region of Competence. KNORA‑U/E use local accuracy oracles \cite{ko2008dynamic}, META‑DES predicts competence via a meta‑learner trained on multiple meta‑features \cite{cruz2015meta}, DES‑P selects classifiers that perform locally better than a random classifier \cite{woloszynski2012measure}, and recent methods leverage graph neural networks \cite{Souza2024} or fuzzy neural networks \cite{davtalab2024scalable} to model more flexible competence patterns.

The key assumption in DES is that a classifier that succeeds on samples similar to the query will also succeed on the query itself \cite{woods1997combination}. By focusing on these local competence estimates, DES handles hard cases, such as samples near decision boundaries, more effectively than static models \cite{cruz2018dynamic}. However, to the best of our knowledge, DES work assumes a single feature space. DRES addresses this gap with a two‑stage framework that first picks the best representation per sample based on hardness measures, then applies competence estimation heuristics and classifier selection from DES techniques within the selected representation space. 

\begin{figure*}[hbt]
    \centering
    \includegraphics[width=0.95\textwidth]{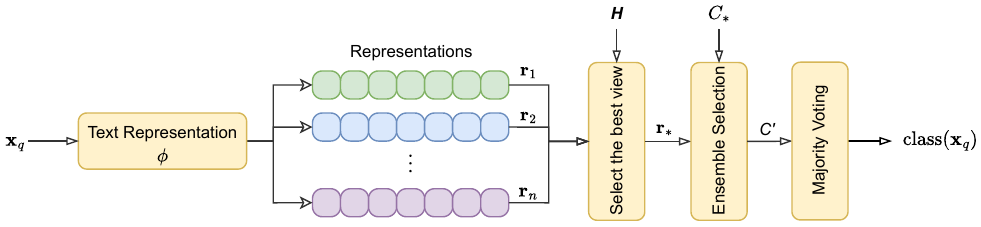}
    \caption{Dynamic Representation and Ensemble Selection (DRES) generalization phase. $\mathbf{x}_q$ is an unknown sample, $\mathbf{r}_j = \phi_j(\mathbf{x}_q)$ is the text representation for $\mathbf{x}_q$, $\mathbf{H}$ is the instance hardness matrix, $\mathbf{r}_*$ is the best text representation associated with the lowest hardness score, $C_*$ is a pool of classifiers trained using the representation $\mathbf{R}_*$, and $C' \subset C_*$ is the most competent subset of classifiers to predict the class of $\mathbf{x}_q$.}
    \label{fig:generalization}
\end{figure*}

\section{Dynamic Representation and Ensemble Selection (DRES)}
\label{sec:Description}

DRES tackles the limitations of fixed fusion and static ensembles by adapting both representation and classifier selection to each input. During training, we compute an instance hardness profile for each sample under every text representation. At inference, we introduce a novel test‐time instance hardness estimation by relating a new article to its neighbors’ training‐time hardness to choose the easiest representation, then dynamically select the most competent classifiers in that space to make the final decision.

\subsection{Training Phase}\label{sec:Training Phase}

In the training phase, we embed the training set $\mathcal{T} = \{\mathbf{x}_1, \mathbf{x}_2, \dots, \mathbf{x}_{|\mathcal{T}|}\}$
into $n$ representation matrices $\mathcal{R} = \{\mathbf{R}_1, \mathbf{R}_2, \ldots, \mathbf{R}_n\}$ using text representation algorithms $\phi = \{\phi_1, \phi_2, \ldots, \phi_n\}$. That is, $\mathbf{R}_j = \phi_j(\mathcal{T})$, and each $\mathbf{R}_j \in \mathbb{R}^{|\mathcal{T}| \times d_j}$ has its own dimensionality $d_j$ (e.g., 768 for BERT and 4096 for Mistral). Afterward, two processes occur: instance hardness calculation and classifier training.

\noindent \textbf{Instance Hardness Calculation.} The instance hardness (IH) of each sample in $\mathcal{T}$ is calculated to compose $\mathbf{H}\in\mathbb{R}^{|\mathcal{T}|\times n}$. Each element $h_{ij} \in \mathbf{H}$ denotes the hardness of instance $\mathbf{x}_i$ when encoded using the $\mathbf{R}_j$ representation. Among the instance hardness metrics, we selected the kDN metric due to its high correlation with classification errors~\cite{smith2014instance, paiva2022relating}.
Equation~\ref{eq:IH} shows how $h_{ij}$ is calculated.

\begin{equation} \label{eq:IH}
\begin{split}
h_{ij} & = \text{kDN}(\mathbf{x}_i, \mathbf{R}_j, k) \\
 & = \frac{\left| \left\{ (\mathbf{x}_l, \mathbf{y}_l) \in \mathcal{N}_k(\mathbf{x}_i;\mathbf{R}_j) : \mathbf{y}_l \neq \mathbf{y}_i \right\} \right|}{k}
\end{split}
\end{equation}

\noindent where, $\mathcal{N}_k(\mathbf{x}_i; \mathbf{R}_j)$ denotes the set of the \(k\)-nearest neighbors of $\mathbf{x}_i$ in the feature space defined by the representation $\mathbf{R}_j$. In other words, kDN counts the number of neighboring instances whose labels differ from that of $\mathbf{x}_i$. Thus, a higher $h$ value indicates that the instance resides in a region of class overlap and is, therefore, difficult to classify correctly.

\noindent \textbf{Classifier Training.} For each representation $\mathbf{R}_j \in \mathcal{R}$, we generate a set $C_j = \{ c_{j1}, c_{j2}, \ldots, c_{jm} \}$ of $m$ classifiers, each trained with a different learning algorithm. This results in $n$ pool of classifiers $\mathcal{C} = \{ C_1, C_2, \ldots, C_n \}$, where each $C_j$ corresponds to models trained on the same representation $\mathbf{R}_j$. In total, the framework produces $n \times m$ classifiers spanning all combinations of representation and learning algorithms.

\subsection{Generalization Phase}\label{sec:Generalization Phase}

Given a new news article as input ($\mathbf{x}_q$), the generalization phase consists of three main steps (Figure~\ref{fig:generalization}):

\noindent \textbf{Text Representation.} 
$\mathbf{x}_q$ is transformed into $n$ different representations, denoted as $\{ \mathbf{r}_1, \mathbf{r}_2, \dots, \mathbf{r}_n \}$, where $\mathbf{r}_j = \phi_j(\mathbf{x}_q)$. Our proposed DRES framework is agnostic regarding the number and type of text representations used.
    
\noindent \textbf{Representation Selection.} We estimate the difficulty of predicting the sample query $\mathbf{x}_q$ across different feature spaces ($\{\mathbf{r}_1, \mathbf{r}_2, \ldots, \mathbf{r}_n$\}, as shown in Figure~\ref{fig:generalization}). The goal is to determine how hard it is to predict $\mathbf{x}_q$ based on each representation and use this information to guide the selection of the optimal feature space. The central hypothesis is that focusing on the classifiers trained using the easiest representation will lead to better prediction outcomes for $\mathbf{x}_q$.
    
\begin{figure}[hbt]
    \centering
    \includegraphics[width=0.38\textwidth]{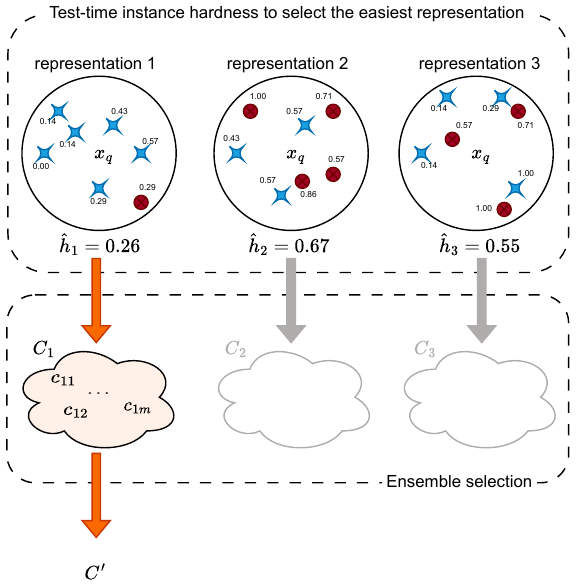}
        \caption{Test-time instance hardness calculation and Ensemble selection. The input text $\mathbf{x}_q$ and its neighbors in three different representation spaces. Instances from two classes (blue and red) are shown associated with their instance hardness values extracted from $\mathbf{H}$. $\hat{h}_j$ represents the instance hardness of $\mathbf{x}_q$ under representation $\mathbf{R}_j$. $C_j$ is the pool of classifiers trained with the same representation and $C' \subset C_1$ is the most competent subset of classifiers to classify $\mathbf{x}_q$.}
    \label{fig:roc-ih-views}
\end{figure}

Since kDN is a supervised metric and we lack access to labels during generalization, we propose an unsupervised estimation of the instance hardness score for the new query sample, called {\it test-time instance hardness}. It works as follows: for each representation $\mathbf{r}_j$, we identify its $k$-nearest neighbors in its corresponding training data embeddings matrix $\mathbf{R}_j$ using the $k$-Nearest Neighbors algorithm. We then estimate its test-time instance hardness score $\hat{h}_j$ by averaging the precomputed instance hardness scores $h_{lj} \in \mathbf{H}$ of these neighbors:
   
\vspace{-1em}

   \begin{equation}
   \label{eq:estimated_hardness}
   \hat{h}_j = \frac{1}{k} \sum_{l=1}^{k} h_{lj}
   \end{equation}

\noindent where $h_{lj}$ is the precomputed hardness score of the $l$-th nearest neighbor in the representation $\mathbf{R}_j$~(Eq.~\ref{eq:IH}). This process is shown in the upper part of Figure~\ref{fig:roc-ih-views}. As an example, for the representation~1, we first select the seven neighbors of $\mathbf{x}_q$ in $\mathcal{T}$ (as shown in the leftmost figure), and their instance values are retrieved from the matrix $\mathbf{H}$ (these values are shown close to each instance). After, the hardness of $\mathbf{x}_q$ is calculated using Eq.~\ref{eq:estimated_hardness}; so, $\hat{h}_1 = 0.26$. The same procedure is performed for all representations.

Collecting these estimated hardness scores produces a vector $\mathbf{\hat{h}} = \{ \hat{h}_1, \hat{h}_2, \ldots, \hat{h}_n \}$, where each element corresponds to the test-time instance hardness for a specific text representation. Then, the text representation with the lowest estimated instance hardness score (Eq.~\ref{eq:viewselection}) is selected. When multiple representations share the same minimal hardness score $\hat{h}_j$, we select the one with the lowest average instance hardness across the entire training set.

\vspace{-1em}

   \begin{equation}
   \label{eq:viewselection}    
   \mathbf{r}_* = \arg\min_{\mathbf{r}_j} \hat{h}_j.
   \end{equation}

In the example (Figure~\ref{fig:roc-ih-views}), representation~1 is claimed as the best feature space to classify $\mathbf{x}_q$ since it lies in a region where its neighbors are more prone to be correctly classified given their kDN. So, $\mathbf{r}_*$ is defined as representation 1 ($\mathbf{r}_1$).

\noindent \textbf{Ensemble Selection.} After the easiest representation $\mathbf{r}_*$ is chosen for query $\mathbf{x}_q$, the method proceeds to select classifiers from the pool $C_*$ trained on $\mathbf{r}_*$. A dynamic selection mechanism identifies the subset of $C_*$ most competent for $\mathbf{x}_q$, and the final prediction is produced via majority vote.

Thus, as shown in the lower part of Figure~\ref{fig:roc-ih-views}, $C_1 \subset \mathcal{C}$ is selected since its classifiers were trained on the text representation previously chosen (representation~1). Given that not all classifiers $c \in C_1$ are competent to classify $\mathbf{x}_q$, only a subset of classifiers $C' \subset C_1$ is selected. This subset selection is performed by dynamic ensemble selection methods~\cite{cruz2018dynamic}, which have the advantage of addressing the classification task in a customized way since the subset $C'$ depends on the query instance under evaluation. The proposed framework is flexible and can incorporate any dynamic selection algorithms, such as META-DES~\cite{cruz2015meta} or KNORA-E~\cite{ko2008dynamic}. The predictions of the selected ensemble ($C'$) are then combined using the Majority Vote rule to produce the final classification.


\section{Experimental Setting}

\noindent \textbf{Datasets.} We evaluate DRES on three standard fake news detection benchmarks: (1) Liar \citep{wang2017liar} (12.8K statement collected from  PolitiFact, 6 truthfulness categories), (2) COVID \citep{patwa2021fighting} (10.7K fact-checked tweets about the pandemic), and (3) GM \citep{mcintire} (14.1K news articles from reputable sources like the New York Times and Wall Street Journal). These datasets represent diverse fake news scenarios, from political claims to health misinformation. Detailed information on these datasets can be found in Table~\ref{tab:datasets}.

\begin{table*}[h]
\centering
\caption[Benchmark datasets]{Main characteristics of the dataset used in this study.}
\scalebox{0.6}{
\begin{tabular}{lllllcl}
\hline
\multicolumn{1}{l}{\textbf{Dataset}} &
  \multicolumn{1}{l}{\textbf{Domain}} &
  \multicolumn{1}{l}{\textbf{Media}} &
  \multicolumn{1}{l}{\textbf{Fact-checking}} &
  \multicolumn{1}{l}{\textbf{Size}} &
  \multicolumn{1}{l}{\textbf{No.Class}} &
  \multicolumn{1}{l}{\textbf{Class distribution}} \\ \hline
\multicolumn{1}{l}{ {Liar \cite{wang2017liar}}} &
  \multicolumn{1}{l}{ {Politics}} &
  \multicolumn{1}{l}{ {Mainstream media}} &
  \multicolumn{1}{l}{ {Editors \& journalists}} &
  \multicolumn{1}{l}{ {12836}} &
  \multicolumn{1}{c}{ {6}} &
  \multicolumn{1}{l}{(1050, 2511, 2108, 2638, 2466, 2063)} \\ 
\multicolumn{1}{l}{ {George McIntire \cite{mcintire}}} &
  \multicolumn{1}{l}{ {Business,Technology, etc.}} &
  \multicolumn{1}{l}{ {Mainstream media}} &
  \multicolumn{1}{l}{ {journalists}} &
  \multicolumn{1}{l}{ {11000}} &
  \multicolumn{1}{c}{ {2}} &
  \multicolumn{1}{l}{(3151, 3159)} \\ 
\multicolumn{1}{l}{ {Covid \cite{patwa2021fighting}}} &
  \multicolumn{1}{l}{ {Covid-19 \& Health}} &
  \multicolumn{1}{l}{ {Twitter}} &
  \multicolumn{1}{l}{ {Fact-checking websites}} &
  \multicolumn{1}{l}{ {10700}} &
  \multicolumn{1}{c}{ {2}} &
  \multicolumn{1}{l}{(5100, 5600)} \\ \hline

\end{tabular}
}
\label{tab:datasets}
\end{table*}

\noindent \textbf{Models.} We employ a comprehensive set of $m=10$ classification algorithms spanning traditional machine learning (Support Vector Machines, Logistic Regression, K-Nearest Neighbors, Naive Bayes, Multi-layer Perceptron, Random Forest, AdaBoost, XGBoost) and deep learning approaches (Convolutional Neural Networks, Bidirectional LSTMs). These are combined with $n=14$ feature representations, including context-independent models (Word2Vec \cite{mikolov2013distributed}, GloVe \cite{pennington2014glove}, FastText \cite{bojanowski2017enriching}) and context-dependent transformers (BERT \cite{devlin2018bert}, DistilBERT \cite{sanh2019distilbert}, ALBERT \cite{lan2019albert}, RoBERTa \cite{liu2019roberta}, BART \cite{lewis2019bart}, ELECTRA \cite{clark2020electra}, XLNet \cite{yang2019xlnet}, Falcon \cite{almazrouei2023falcon}, ELMo \cite{https://doi.org/10.48550/arxiv.1802.05365}, LLaMA3 \cite{grattafiori2024llama}, and Mistral \cite{jiang2023mistral7b}). Detailed information on these representations can be found in Appendix~\ref{app:representations}.

We use 5-fold cross-validation with hyperparameter tuning (see Appendices~\ref{app:Grid_hyperparameters_classic}~and~\ref{app:Grid_hyperparameters_deep}), considering F1 scores to account for class imbalance.

\noindent \textbf{Baselines.} We evaluate DRES against three static ensemble baselines representing distinct static selection approaches in order to test our hypothesis that dynamic selection of both representations and classifiers outperforms static ensemble strategies:
\setlist{nolistsep}
\begin{itemize}[noitemsep]
\item \textbf{Group A:} 10 pools of the same classifier type across all feature representations (10 single classifiers $\times$ 14 representations).
\item \textbf{Group B:} 14 pools of diverse classifiers per text representation (10 classifiers $\times$ 14 single representation)
\item \textbf{Group C:} Full combination of all 140 classifiers (10 classifiers $\times$  14 representations).
\end{itemize}

The classifiers in each group are combined using stacked generalization~\cite{wolpert1992stacked} with logistic regression as the meta-classifier, as in~\cite{cruz2022selecting}.  

In the DRES method, instance hardness scores were calculated using $k$NN ($k = 5$) to define the region of competence (RoC), i.e., the local neighborhood of the test instance in the validation dataset. A detailed analysis of different \(k\)-values is provided in Appendix~\ref{sec:khardness}. We adopted three dynamic ensemble selection techniques, KNORA-E, DES-P, and META-DES, which were adapted from the DESlib library~\citep{JMLR:v21:18-144} to support multiple text representations. These methods were chosen because they rely on neighborhood-based competence estimation, aligning with our test-time instance hardness approach. 

\section{Results}
\label{sec:results}

\subsection{Comparison with static ensemble}
\label{sec:comparision-with-static-ensemble}

Table~\ref{tab:resultGroups} reports DRES’s performance against the baseline Groups A, B, and C using macro F1. In this table, only the best results for Groups A and B are reported; the complete results and additional metrics are in Appendices~\ref{app:otherperf} and~\ref{app:static-results}.

\begin{table}[h]
\centering
\caption{F1-score per dataset for DRES and baseline models. The absolute best results per dataset are in bold, and the top methods that are statistically equivalent are marked with an asterisk.}
\resizebox{0.48\textwidth}{!}{
\begin{tabular}{lrrr}
\cline{2-4}
& \multicolumn{3}{c}{\textbf{Dataset}} \\ \hline
\textbf{Method}             & \textbf{Liar} & \textbf{COVID} & \textbf{GM} \\ \hline

MLP (Best Group A)        & 0.250 (0.002)   & 0.950 (0.002) & 0.950 (0.003) \\ \hline
Mistral (Best Group B)    & 0.260 (0.002)   & 0.943 (0.003) & 0.951 (0.002) \\ \hline
Group C              & 0.243 (0.003)  & 0.941 (0.003)  & 0.950 (0.002)       \\ \hline
DRES (KNORA-E)       & 0.371 (0.003)  & \textbf{0.973 (0.002)*}  & \textbf{0.986 (0.002)*}       \\ 
DRES (META-DES)      & 0.367 (0.002)         & 0.972 (0.002)*  & 0.980 (0.005)      \\
DRES (DES-P)         & \textbf{0.385 (0.003)*} & 0.972 (0.003)*   & 0.984 (0.002)*      \\
\hline
\end{tabular}
}
\label{tab:resultGroups}
\end{table}

DRES outperforms static strategies by a clear margin for all datasets, independent of the dynamic selection classifier used, as statistically corroborated by the repeated measures ANOVA with Tukey's post-hoc test with a 95\% confidence interval. The tests for all datasets revealed statistically significant differences (p-value $<$ 0.0001), demonstrating the superiority of our method. In Table~\ref{tab:resultGroups}, the methods with the absolute best results are highlighted in bold, and the top methods that are statistically equivalent are marked with an asterisk for easier identification. For the Liar dataset, DRES with DES-P presented the best performance, while for the COVID dataset, the performance differences among DRES methods (KNORA-E, META-DES, and DES-P) were not statistically significant. In turn, DRES with KNORA-E and DES-P presented better results for the GM dataset, with no significant difference between them.

Group B, a homogeneous ensemble that uses only one text representation for all classifiers, fails when the representation is unsuitable for specific instances. 
Group A, limited to a single classifier learning algorithm, cannot adapt across heterogeneous views. Even combining all models and representations (Group C) does not overcome these fundamental shortcomings, yielding consistently lower scores. 
DRES addresses these issues via a dynamic two-stage selection. First, it selects the most appropriate feature representation (i.e., the one deemed easiest for classification for the given instance), and then it dynamically identifies locally competent classifiers trained over the selected text representation. This adaptive mechanism is essential for robust performance in complex fake news detection tasks.


\subsection{Comparison with state-of-the-art models}
\label{comparisionstateoftheart}

\begin{table}[h]
\caption{Liar dataset: DRES versus state-of-the-art methods. The best results are in bold.}
\centering
\resizebox{0.45\textwidth}{!}{%
\begin{tabular}{lc}
\hline
\textbf{Method}                                        &\textbf{F1-Score} \\ \hline
LSTM (GloVe) \cite{rashkin2017truth}                   & 0.210 \\
Hybrid CNN (Word2vec) \cite{wang2017liar}              & 0.274 \\
Logistic Regression (GloVe) \cite{alhindi2018your}     & 0.250 \\
CNN (Word2Vec) \cite{girgis2018deep}                   & 0.270  \\
MMFD (LSTM + Word2vec) \cite{karimi2018multi}          & 0.290 \\
Multi-view autoencoder~\cite{PCC2025}                  & 0.253\\
\hline
DRES (KNORA-E)                                 & \textbf{0.371}        \\ 
DRES (META-DES)                                & \textbf{0.367}        \\
DRES (DES-P)                                   & \textbf{0.385}  \\
\hline

\end{tabular}
}
\label{tab:Liarbenchmark}
\end{table}

\begin{table}[ht]
\caption{COVID dataset: DRES versus state-of-the-art methods. The best results are in bold.}
\centering
\resizebox{0.45\textwidth}{!}{%
\begin{tabular}{lc}
\hline
\textbf{Method}                                           &\textbf{F1-Score} \\ \hline
BiGRU-Attention (BERT) \cite{karnyoto2022transfer}        & 0.919 \\
SVM (TFIDF) \cite{patwa2021fighting}                      & 0.933 \\
BERT (End-to-End) \cite{10074216}                         & 0.944  \\
\hline
DRES (KNORA-E)                     & \textbf{0.973}        \\ 
DRES (META-DES)                    & \textbf{0.972}        \\
DRES (DES-P)                       & \textbf{0.972}     \\        
\hline
\end{tabular}
}
\label{tab:Covidbenchmark}
\end{table}

\begin{table}[h]
\caption{GM dataset: DRES versus state-of-the-art methods. The best results are in bold.}
\centering
\resizebox{0.43\textwidth}{!}{%
\begin{tabular}{lc}
\hline
\textbf{Method}                                              &\textbf{F1-Score} \\ \hline
HDSF (Word2vec) \cite{karimi2019learning}                    & 0.822 \\
XGBoost (Word2vec)  \cite{reddy2020text}                     & 0.860 \\
XGBoost (GloVe) \cite{bali2019comparative}                   & 0.873  \\
Naive Bayes (TFIDF)  \cite{agudelo2018raising}                                & 0.881   \\
Ensemble Learning \cite{elsaeed2021detecting}      & 0.946  \\
BiLSTM (GloVe)  \cite{SASTRAWAN2022396}                                & 0.948  \\
\hline
DRES (KNORA-E)                     & \textbf{0.986}        \\ 
DRES (META-DES)                    & \textbf{0.980}        \\
DRES (DES-P)                       & \textbf{0.984}  \\  
\hline
\end{tabular}
}
\label{tab:GMbenchmark}
\end{table}

\begin{table*}[tbh]
\caption{Impact of DRES components. Blue values show improvements (in percentage points, pp) over the baseline (Group C). Oracle configurations (bottom) represent the theoretical upper bounds.}
\centering
\scalebox{0.8}{
\begin{tabular}{llll}
\hline
\textbf{Method} & \textbf{Liar} & \textbf{COVID} & \textbf{GM} \\ \hline \hline
\textit{Group C (no selection)} & 0.243 & 0.941 & 0.950 \\
\midrule

Dynamic Ensemble selection only & 0.299\,\textcolor{blue}{(+\,5.6 pp)} & 0.948\,\textcolor{blue}{(+\,0.7 pp)} & 0.963\,\textcolor{blue}{(+\,1.3 pp)} \\

Representation selection only & 0.345\,\textcolor{blue}{(+\,10.2 pp)} & 0.955\,\textcolor{blue}{(+\,1.4 pp)} & 0.971\,\textcolor{blue}{(+\,2.1 pp)} \\

DRES (KNORA-E) & \textbf{0.371}\,\textcolor{blue}{(+\,12.8 pp)} & \textbf{0.961}\,\textcolor{blue}{(+\,2.0 pp)} & \textbf{0.982}\,\textcolor{blue}{(+\,3.2 pp)} \\

\midrule
\midrule
DRES + Oracle (representation) & 0.832 & 0.996 & 0.999 \\
DRES + Oracle (representation + classifier) & 0.995 & 0.999 & 1.000 \\ \hline
\end{tabular}
}

\label{tab:dmes_ablation}
\end{table*}

Tables~\ref{tab:Liarbenchmark},~\ref{tab:Covidbenchmark}, and~\ref{tab:GMbenchmark} compare DRES with state-of-the-art models for the Liar, COVID, and GM datasets, respectively, focusing on text-based fake news detection. DRES achieves the highest macro F1 score on the LIAR dataset (Table~\ref{tab:Liarbenchmark}) using only textual features. It is important to note that the Liar dataset contains additional metadata about the speaker's profile and justification in addition to the news statements that were not used in this analysis, as these auxiliary features are often unavailable or inconsistent across datasets.

For the COVID (Table~\ref{tab:Covidbenchmark}) and GM (Table~\ref{tab:GMbenchmark}) datasets, the proposed DRES model achieved the top-ranking with an F1-score of 0.971 (with DES-P) and 0.982 (with KNORA-E), respectively. These results highlight our model's capability in another domain (medical information), where reliability is of the highest importance.
In all three scenarios, DRES obtained better results than SOTA, regardless of the selected DES algorithm used. Additional experiments with fine-tuned large LLMs are reported in Appendix~\ref{app:finetuning}, confirming that DRES remains competitive even against stronger contextual baselines.

\subsection{Ablation Study}
\label{Ablation Study}

In this section, we evaluate the contribution of view selection and classifier selection phases within the DRES framework by comparing three variants: (i)~only view selection and majority voting over classifiers, (ii)~only classifier selection on a fixed view (Mistral), and (iii)~the DRES system (Table~\ref{tab:dmes_ablation}). 

The proposed dynamic view selection scheme alone significantly improves the baseline Group C, with gains of +10.2, +1.4, and +2.1 percentage points on the Liar, Covid, and GM datasets, respectively. These improvements show that dynamically selecting representations helps address textual ambiguity, particularly for the liar dataset (6 classes). 
Moreover, when combined with KNORA-E’s dynamic classifier selection, DRES consistently achieves further improvements across all datasets. This improvement is more evident for the Liar dataset (+5.6 percentage points), which can be explained by the fact that this is a harder dataset (full hardness heatmap for all datasets are presented in Appendix~\ref{app:hardnessheatmap}).
These results highlight the effectiveness of a two-stage selection approach, where dynamic view and classifier selection work together to improve overall performance, particularly in the presence of residual ambiguity (i.e.,  classification uncertainty that persists after representation selection due to overlap).

\begin{figure*}[tbh]
    \centering
        \includegraphics[width=0.8\textwidth]{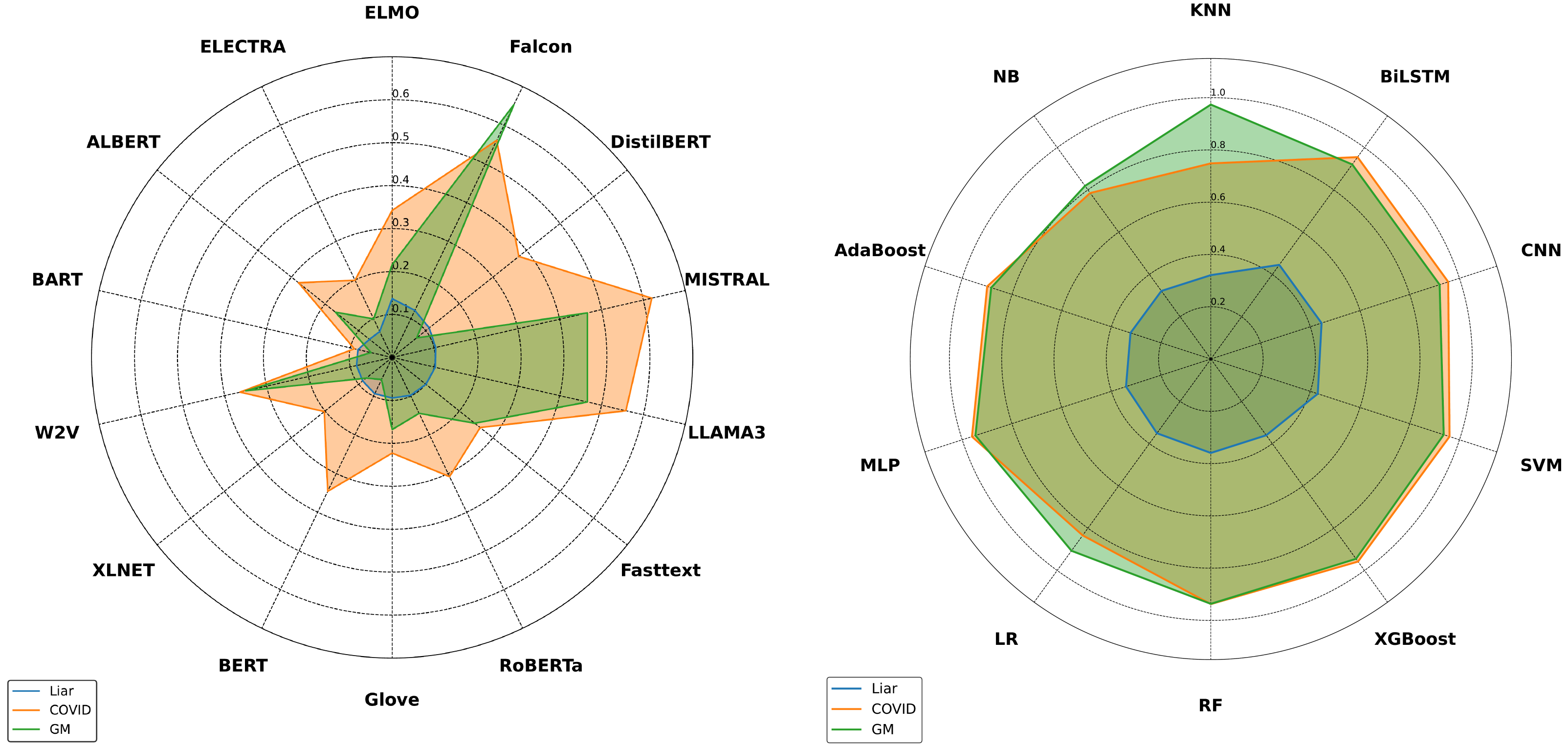}
     \caption{Frequency of selected views (left) and classifiers (right) for each dataset using the KNORA-E dynamic ensemble selection method.}
     \label{fig:Frequency-view}
\end{figure*}

The last two lines of Table~\ref{tab:dmes_ablation} show the Oracle results for i)~the selection of the best representation and ii)~the selection of both best representation and best classifier per new instance. The Oracle represents a theoretical upper bound since it always selects the best representation and classifier per query instance. First, perfect representation selection could lead to a +58.9 percentage points improvement over the baseline on Liar, pointing to significant room for improvement in representation selection. Second, achieving nearly perfect accuracy with complete Oracle settings (99.5–100\%) validates DRES’s architecture when both phases work optimally. Hence, the gap with the Oracle suggests that better instance hardness estimators should be investigated in order to improve the representation selection stage.

\subsection{Selection Analysis}
\label{Selection Analysis}

DRES framework's instance-specific selection of classifier-view pairs is analyzed in Figure~\ref{fig:Frequency-view} through complementary radar plots showing i)~representation selection (left) and ii)~classifier selection frequencies (right) across datasets, using KNORA-E for dynamic ensemble selection. The results reveal significant diversity among the chosen views and classifiers. While significant diversity in selected combinations confirms the value of the multi-stage dynamic selection approach, the underutilization of specific components may suggest they could be pruned without affecting overall performance.

\subsection{Instance hardness analysis}

\begin{figure}[htb]
    \centering
       \includegraphics[width=.48\textwidth]{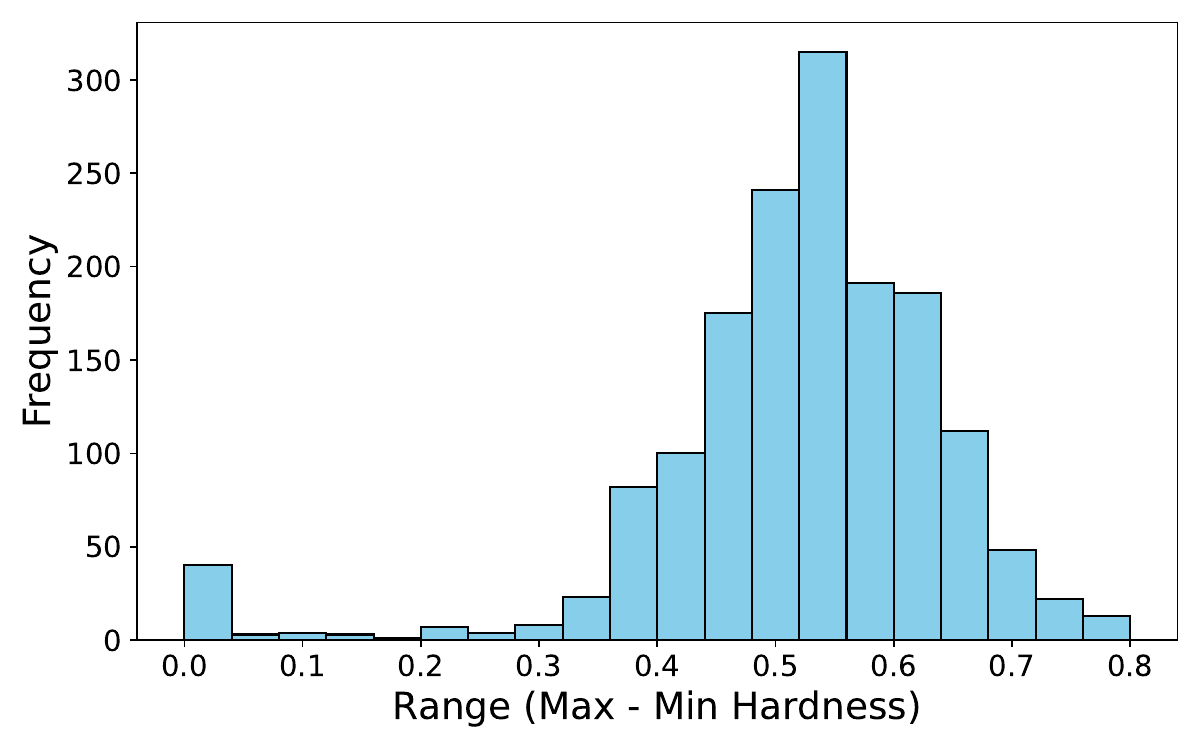}
         \caption{Distribution of the range (max-min) instance hardness computed for the GM dataset.}
        \label{fig:statsGM}
\end{figure}

\begin{figure}[h]
    \centering
       \includegraphics[width=0.9\linewidth]{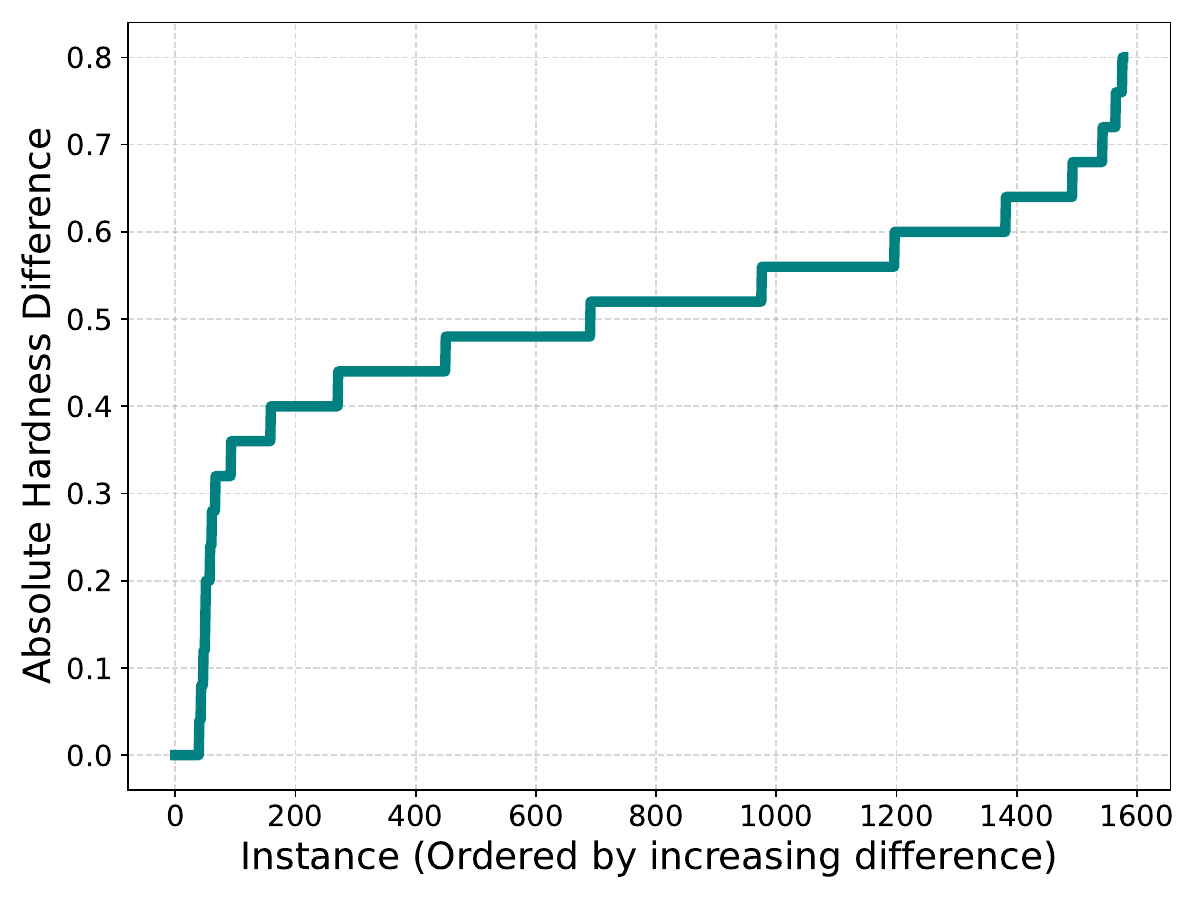}
        \caption{Cumulative distribution of the difference between maximum and minimum instance hardness values for the GM dataset.}
      \label{fig:ascendingGM}
\end{figure}

\label{sec:hardnessVariation}
We investigate how classification difficulty varies across representations by analyzing instance-level hardness scores on the GM dataset. Figure~\ref{fig:statsGM} shows the range of hardness scores across views for each instance, computed as the difference between the maximum and minimum values. Over 50\% of instances show a range above 0.5 and 25\% exceed 0.7, indicating substantial disagreement between representations. Figure~\ref{fig:ascendingGM} further confirms that this variation is widespread and not limited to a few outliers. These findings highlight the limitations of relying on a single representation or a fixed set, as they may perform inconsistently across different inputs. Our results reinforce the motivation for a dynamic selection strategy based on test-time hardness estimation. Full results, including standard deviation and coefficient of variation analyses, as well as figures for the Liar and COVID datasets, are provided in Appendix~\ref{app:variationrepresentation}.

\section{Conclusion}

In this paper, we introduced DRES, a fake news detection model that dynamically selects the most suitable feature representations and classifier ensembles for each input text by leveraging instance hardness analysis and dynamic classifier selection. We evaluated our approach on three diverse datasets. We found that it outperformed state-of-the-art methods and baseline models, especially when only textual information was considered, which is the most prevalent and accessible form of information in fake news detection. In addition, several ablation studies demonstrate that one must consider a dynamic system in both view and classifier selection in order to obtain higher accuracy. Future work will involve adapting DRES to integrate multiple information sources, such as network propagation and user behavior data, as well as investigating new mechanisms for test-time hardness calculation.

\section{Limitations}

While DRES advances text-based fake news detection through dynamic representation and classifier selection, three key limitations merit discussion. First, our evaluation focuses on English datasets (Liar, COVID, GM), which limits insights into low-resource languages like Urdu or Arabic, which pose unique challenges~\cite{harris2025benchmarking,albtoush2025fake}. Furthermore, we did not test DRES on synthetic datasets containing LLM-generated misinformation, despite evidence that detectors often misclassify such content due to linguistic biases~\cite{su2023fake}. Future work should validate DRES against adversarial LLM outputs to assess its robustness in diverse scenarios. 

 Second, DRES currently selects only one representation per instance, potentially overlooking complementary signals from views with similar hardness levels. Additionally, our hardness estimation relies solely on the kDN metric, while alternative measures~\cite{ethayarajh22a} might capture different aspects of instance difficulty. Future work should explore multi-representation fusion strategies and hardness metric ensembles to better handle borderline cases where multiple representations appear equally viable.

Third, while DRES dynamically selects representations and classifiers, its training phase incurs upfront costs from computing instance hardness across multiple representations. As shown in Figure~\ref{fig:Frequency-view}, certain representations like BART are selected in <10\% of cases across datasets, suggesting potential redundancy. Pruning such infrequently used representations guided by data complexity~\cite{cook2025no} or embedding diversity could simplify the system without compromising performance. 

Lastly, it is important to acknowledge the possibility of training data contamination for recently released language models such as LLaMA3 and Mistral. These models were made available after the release of the datasets used in this study (LIAR, COVID, GM), suggesting that portions of these datasets may have been included in their pretraining corpora. This raises concerns about potential memorization effects, particularly in fine-tuning scenario. However, this issue does not compromise the core findings of our work that is: instance-level hardness varies across representation, and dynamic selection of both representation and classifier improves performance.

\bibliography{paper}


\appendix

\section{Supplementary details on representations}
\label{app:representations}

Table~\ref{tab:embedding_models} provides a comprehensive list of the models, their implementation sources (e.g., HuggingFace, Zeugma, AllenNLP), and embedding dimensions used in our experiments. This includes both classical word embedding methods (e.g., Word2Vec, GloVe) and modern pretrained language models (e.g., BERT, LLaMA, Mistral). 

\begin{table}[H]
\centering
\caption{Text representation techniques used in this study.}
\label{tab:embedding_models}
\resizebox{0.48\textwidth}{!}{ 
\begin{tabular}{lllr}
\hline
\textbf{Model} & \textbf{Version/Name} & \textbf{Source} & \textbf{Dimension} \\
\hline
Word2Vec & word2vec & Zeugma & 300 \\
GloVe & glove & Zeugma & 300 \\
FastText & fasttext & Zeugma & 300 \\
ELMO & allenai/elmo & AllenNLP & 1024 \\
BERT & bert-base-uncased & HuggingFace & 768 \\
DistilBERT & distilbert-base-uncased & HuggingFace & 768 \\
ALBERT & albert-base-v2 & HuggingFace & 768 \\
RoBERTa & roberta-base & HuggingFace & 768 \\
BART & facebook/bart-base & HuggingFace & 768 \\
ELECTRA & google/electra-base-discriminator & HuggingFace & 768 \\
XLNet & xlnet-base-cased & HuggingFace & 768 \\
LLaMA & CodeLlama-7b & HuggingFace & 4096 \\
Falcon & falcon & HuggingFace & 2048 \\
LLaMA3 & Llama3.2-1B & HuggingFace & 2048 \\
Mistral & mistral-7B-v0.1 & HuggingFace & 4096 \\
\hline
\end{tabular}
}
\end{table}

\section{Hyperparameters for classical and ensemble models}
\label{app:Grid_hyperparameters_classic}

\begin{table}[H]
\centering
\caption{Hyperparameters considered for machine learning and ensemble models.}
\resizebox{0.48\textwidth}{!}{%
		\begin{tabular}{|l|l|}
			\hline
			\textbf{Method} & \textbf{Hyperparameters}
			\\ \hline
			\textbf{SVM} & \begin{tabular}[c]{@{}l@{}}Kernel:['rbf']\\ Gamma: [1, 0.1, 0.01, 0.001, 0.0001]\\ C: [0.1, 1, 10, 100, 1000]\end{tabular} \\ \hline
			
			\textbf{LR} & \begin{tabular}[c]{@{}l@{}}solver: ['liblinear']\\ penalty: ['none', 'l1', 'l2', 'elasticnet']\\ C: [0.01, 0.1, 1, 10, 100]\end{tabular} \\ \hline
			
			\textbf{NB} & \begin{tabular}[c]{@{}l@{}}alpha: [0.1, 0.5, 1]\\ fit\_prior: [False, True]\end{tabular} \\ \hline
			
			\textbf{KNN} & n\_neighbors: [1 - 20] \\ \hline
				
			\textbf{RF} & \begin{tabular}[c]{@{}l@{}}bootstrap: [True, False]\\ max\_depth: [5,10, 20, 30, 40, 50]\\ max\_features: ['auto', 'sqrt', 'log2']\\ min\_samples\_leaf: [1, 2, 4]\\ min\_samples\_split: [2, 5, 10]\\ n\_estimators: [200, 400, 600, 800, 1000]\\  criterion: ['gini', 'entropy']\end{tabular} \\ \hline

			\textbf{AdaBoost} & \begin{tabular}[c]{@{}l@{}}n\_estimators: [10, 50, 100, 200, 300, 400, 500, 1000]\\ learning\_rate: [0.001, 0.01, 0.1, 0.2, 0.5]\end{tabular} \\ \hline
			
			\textbf{XGBoost} & \begin{tabular}[c]{@{}l@{}}n\_estimators: [200,300,400,500]\\ max\_features: ['sqrt', 'log2']\\ max\_depth: [4,5,6,7,8]\\ criterion: ['gini', 'entropy']\\ random\_state: [18]\end{tabular} \\ \hline
   
   			\textbf{MLP} & \begin{tabular}[c]{@{}l@{}}Activation function: [ReLU, logistic] \\ solver: [Adam, lbfgs]\end{tabular} \\ \hline
			
		\end{tabular}%
		}
	\label{tab:hyperparameter_grid-1}
\end{table}

Table~\ref{tab:hyperparameter_grid-1} outlines the grid search ranges used to tune classical and ensemble models. For SVM, we tuned the kernel coefficient (gamma) and regularization strength (C), which are known to have the most significant impact on model performance. Logistic regression was configured across solvers and regularization types. Naive Bayes used different smoothing values and prior settings. For KNN, we adjusted the number of neighbors. Random Forest, AdaBoost, and XGBoost were tuned by varying tree depths, number of estimators, and split criteria. MLP models were tested with different activation functions and solvers.

\section{Hyperparameters for deep learning models}
\label{app:Grid_hyperparameters_deep}

Table~\ref{tab:hyperparameter_grid-2} shows the tuning setup for CNN and BiLSTM models. Both were trained using the Adam optimizer with a fixed learning rate and dropout. We varied activation functions, batch sizes, and the number of epochs. For BiLSTM, we also included hidden size as a tuning parameter.

\begin{table}[H]
\centering
\caption{Hyperparameters considered for all deep learning models.}
\resizebox{0.36\textwidth}{!}{%
		\begin{tabular}{|l|l|}
			\hline
			\textbf{Method} & \textbf{Setup}
			\\ \hline
			
			\textbf{CNN} & \begin{tabular}[c]{@{}l@{}}Activation function: [sigmoid, ReLU] \\ Batch size: [64, 128, 512] \\ Number of epochs: [5,20,100] \\ Optimizer: [Adam] \\ Learning rate: 0.001 \\ Dropout: 0.2 \end{tabular} \\ \hline
			
			\textbf{BiLSTM} & \begin{tabular}[c]{@{}l@{}}Activation function: [sigmoid, ReLU]\\ Batch size: [64, 128, 512]\\ Number of epochs: [5,20,100] \\ Optimizer: [Adam] \\ Learning rate: 0.001\\ Hidden size: 128\\ Dropout: 0.2\end{tabular} \\ \hline

			\end{tabular}%
	    }
	\label{tab:hyperparameter_grid-2}
\end{table}

\section{The impact of the k-hardness hyperparameter}
\label{sec:khardness}

We conducted additional experiments to assess the impact of the number of neighbors ($k$) used in test-time instance hardness estimation. We evaluated DRES with $k \in \{3, 5, 7, 9, 11, 13\}$. The results are presented in Table~\ref{tab:khardnessresults}.

\begin{table}[ht]
\centering
\caption{Impact of $k$ on DRES performance across datasets.}
\label{tab:khardnessresults}
\resizebox{0.45\textwidth}{!}{%
\begin{tabular}{lcccc}
\toprule
\textbf{Method}        & \textbf{$k$} & \textbf{COVID} & \textbf{GM} & \textbf{Liar} \\
\midrule
DRES (KNORA-E)  & 3         & 0.964 (0.002)          & 0.981 (0.002)       & 0.366 (0.003)         \\
DRES (KNORA-E)  & 5         & 0.973 (0.002)          & 0.986 (0.002)       & 0.371 (0.003)      \\
DRES (KNORA-E)  & 7         & 0.968 (0.007)          & 0.984 (0.005)       & 0.366 (0.005)         \\ 
DRES (KNORA-E)  & 9         & 0.97 (0.001)           & 0.982 (0.004)       & 0.365 (0.004)         \\ 
DRES (KNORA-E)  & 11        & 0.965 (0.002)          & 0.981 (0.006)       & 0.367 (0.003)         \\ 
DRES (KNORA-E)  & 13        & 0.971 (0.002)          & 0.979 (0.001)       & 0.366 (0.002)         \\ \hline
DRES (META-DES) & 3         & 0.969 (0.001)          & 0.971 (0.001)       & 0.362 (0.007)         \\
DRES (META-DES) & 5         & 0.972 (0.002)          & 0.98 (0.005)        & 0.367 (0.002)          \\
DRES (META-DES) & 7         & 0.97 (0.009)           & 0.975 (0.002)       & 0.363 (0.002)        \\ 
DRES (META-DES) & 9         & 0.964 (0.002)          & 0.974 (0.003)       & 0.361 (0.001)        \\ 
DRES (META-DES) & 11        & 0.97 (0.001)           & 0.977 (0.001)       & 0.362 (0.002)        \\ 
DRES (META-DES) & 13        & 0.966 (0.001)          & 0.979 (0.006)       & 0.363 (0.001)        \\ \hline
DRES (DES-P)    & 3         & 0.968 (0.001)          & 0.976 (0.003)       & 0.381 (0.004)        \\
DRES (DES-P)    & 5         & 0.972 (0.003)          & 0.984 (0.002)       & 0.385 (0.003)          \\
DRES (DES-P)    & 7         & 0.969 (0.001)          & 0.981 (0.001)       & 0.381 (0.003)         \\
DRES (DES-P)    & 9         & 0.971 (0.004)          & 0.977 (0.002)       & 0.383 (0.002)          \\
DRES (DES-P)    & 11        & 0.968 (0.001)          & 0.976 (0.007)       & 0.38 (0.008)          \\
DRES (DES-P)    & 13        & 0.966 (0.001)          & 0.976 (0.003)       & 0.377 (0.009)         \\
\bottomrule
\end{tabular}
}
\end{table}

The results indicate that DRES is robust to the choice of $k$, with only marginal differences in F1-scores across values. While $k=5$ slightly outperforms other settings, the variation remains within a narrow range, suggesting that the method does not rely heavily on fine-tuning this parameter.


\section{Performance for other metrics}
\label{app:otherperf}

Table~\ref{tab:resultPrecision} reports DRES’s performance for the precision results, while Table~\ref{tab:resultRecall} reports the recall results. 

\begin{table}[h]
\centering
\caption{Precision performance of DRES models across datasets.}
\resizebox{0.48\textwidth}{!}{
\begin{tabular}{lccc}
\cline{2-4}
& \multicolumn{3}{c}{\textbf{Dataset}} \\ \hline
\textbf{Method}      & \textbf{Liar} & \textbf{COVID} & \textbf{GM} \\ \hline

DRES (KNORA-E)       & 0.380 (0.003) & 0.970 (0.002) &  0.985 (0.002)  \\ 
DRES (META-DES)      & 0.377 (0.002) & 0.968 (0.002) & 0.978 (0.004) \\
DRES (DES-P)         & 0.390 (0.003) & 0.970 (0.003) &  0.983 (0.002)  \\
\hline
\end{tabular}
}
\label{tab:resultPrecision}
\end{table}

\begin{table}[h]
\centering
\caption{Recall performance of DRES models across datasets.}
\resizebox{0.48\textwidth}{!}{
\begin{tabular}{lccc}
\cline{2-4}
& \multicolumn{3}{c}{\textbf{Dataset}} \\ \hline
\textbf{Method}      & \textbf{Liar} & \textbf{COVID} & \textbf{GM} \\ \hline

DRES (KNORA-E)       & 0.365 (0.003) & 0.975 (0.002) & 0.988 (0.002)  \\ 
DRES (META-DES)      & 0.358 (0.002) & 0.975 (0.002) & 0.982 (0.006) \\
DRES (DES-P)         & 0.380 (0.003) & 0.974 (0.003) & 0.985 (0.002)   \\
\hline
\end{tabular}
}
\label{tab:resultRecall}
\end{table}


\section{Static ensemble results}
\label{app:static-results}

Table~\ref{tab:resultGroupsAppendix} expands upon the main manuscript's findings by detailing the performance of individual models within Groups A, B, and C across the Liar, COVID, and GM datasets. The results reveal that within each group, ensemble models exhibit similar performance levels, indicating that static combination methods, even when combined with a meta-classifier with stacked generalization (i.e., a learned combination scheme). 

In contrast, the DRES framework, when integrated with dynamic ensemble selection methods such as KNORA-E, META-DES, and DES-P, consistently outperforms the baseline groups across all datasets. Notably, the choice among these dynamic selection techniques results in marginal performance differences, suggesting that the primary advantage arises from the dynamic selection mechanism itself rather than the specific competence estimation heuristics and classifier selection approaches they employ~\cite{cruz2018dynamic}.

\begin{table}[]
\centering
\caption{Results of F1-Score per dataset for DRES and baseline models (Groups A, B, and C).}
\resizebox{0.48\textwidth}{!}{
\begin{tabular}{lrrr}
\cline{2-4}
& \multicolumn{3}{c}{\textbf{Dataset}} \\ \hline
\textbf{Method}             & \textbf{Liar} & \textbf{COVID} & \textbf{GM} \\ \hline
\textbf{LR (Group A)}         & \textbf{0.260} & 0.940           & \textbf{0.950}        \\
\textbf{SVM (Group A)}        & \textbf{0.260} & 0.940           & 0.940        \\
\textbf{KNN (Group A)}        & 0.230          & 0.930           & 0.890        \\
\textbf{NB (Group A)}         & 0.220          & 0.920           & 0.860        \\
\textbf{XGBoost (Group A)}    & 0.250          & 0.940           & \textbf{0.950}        \\
\textbf{RF (Group A)}         & \textbf{0.260} & 0.920           & 0.920        \\
\textbf{AdaBoost (Group A)}   & 0.190          & 0.910 	         & 0.920        \\
\textbf{BiLSTM (Group A)}     & 0.250          & 0.930           & 0.920        \\
\textbf{CNN (Group A)}        & 0.250          & 0.940           & 0.920        \\
\textbf{MLP (Group A)}        & 0.250          & \textbf{0.950}  & \textbf{0.950}        \\ \hline

\textbf{TF (Group B)}         & 0.250            & 0.930           & 0.940        \\
\textbf{TFIDF (Group B)}      & 0.230            & 0.940           & 0.950        \\
\textbf{W2V (Group B)}        & 0.230          & 0.910           & 0.920        \\
\textbf{GloVe (Group B)}      & 0.220          & 0.860           & 0.840        \\
\textbf{FastText (Group B)}   & 0.240          & 0.910           & 0.900        \\
\textbf{ELMO (Group B)}       & 0.240          & 0.910           & 0.920        \\
\textbf{BERT (Group B)}       & 0.230          & 0.910           & 0.820        \\
\textbf{DistilBERT (Group B)} & 0.220          & 0.920           & 0.830        \\
\textbf{RoBERTa (Group B)}    & 0.230          & 0.910           & 0.850        \\
\textbf{ALBERT (Group B)}     & 0.240          & 0.900           & 0.840        \\
\textbf{BART (Group B)}       & 0.220          & 0.880           & 0.850        \\
\textbf{ELECTRA (Group B)}    & 0.230          & 0.910           & 0.850        \\
\textbf{XLNET (Group B)}      & 0.220           & 0.900           & 0.860        \\
\textbf{Falcon (Group B)}     & 0.250          & 0.940           & 0.950        \\
\textbf{LLaMA3 (Group B)}     & \textbf{0.261}         & 0.940           & 0.950        \\
\textbf{Mistral (Group B)}    & 0.260         & \textbf{0.943}           & \textbf{0.951}        \\\hline

\textbf{Group C}              & 0.243          & 0.941           & 0.950        \\ \hline
\textbf{DRES + KNORA-E}       & 0.371          & \textbf{0.973}           & \textbf{0.986}       \\
\textbf{DRES + META-DES}      & 0.367          & 0.972           & 0.980       \\
\textbf{DRES + DES-P}         & \textbf{0.385} & 0.972           & 0.984       \\

\hline
\end{tabular}
}
\label{tab:resultGroupsAppendix}
\end{table}

\section{End-to-end LLM fine-tuned results}
\label{app:finetuning}

\begin{table}[h]
\centering
\caption{End-to-end fine-tuned model results (F1-score (std)) across the LIAR, COVID, and GM datasets. The absolute best results per dataset are in bold, and the top DRES methods that are statistically equivalent are marked with an asterisk.}
\resizebox{0.48\textwidth}{!}{
\begin{tabular}{lrrr}
\cline{2-4}
& \multicolumn{3}{c}{\textbf{Dataset}} \\ \hline
\textbf{Model}             & \textbf{Liar} & \textbf{COVID} & \textbf{GM} \\ \hline
\textbf{ELMO}              & 0.210 (0.001) & 0.935 (0.001) & 0.928 (0.005) \\
\textbf{BERT}              & 0.210 (0.003) & 0.880 (0.002) & 0.880 (0.006) \\
\textbf{DistilBERT}        & 0.220 (0.008) & 0.940 (0.006) & 0.890 (0.003) \\
\textbf{ALBERT}            & 0.180 (0.000) & 0.840 (0.001) & 0.870 (0.001) \\
\textbf{BART}              & 0.200 (0.001) & 0.890 (0.002) & 0.940 (0.001) \\
\textbf{RoBERTa}           & 0.190 (0.010) & 0.850 (0.009) & 0.850 (0.009) \\
\textbf{ELECTRA}           & 0.190 (0.002) & 0.860 (0.004) & 0.890 (0.001) \\
\textbf{XLNET}             & 0.210 (0.003) & 0.860 (0.003) & 0.890 (0.011) \\
\textbf{LLaMA}             & 0.261 (0.003) & 0.941 (0.002) & 0.992 (0.003) \\
\textbf{Falcon}            & 0.258 (0.001) & 0.948 (0.001) & 0.993 (0.000) \\
\textbf{Mistral}           & 0.268 (0.002) & 0.950 (0.002) & 0.994 (0.001) \\
\textbf{LLaMA3}            & 0.270 (0.003) & 0.953 (0.002) & \textbf{0.995 (0.001)} \\ \hline

\textbf{MLP (Group A)}     & 0.250 (0.002) & 0.950 (0.002) & 0.950 (0.003) \\
\textbf{Mistral (Group B)} & 0.260 (0.002) & 0.943 (0.003) & 0.951 (0.002) \\
\textbf{Group C}           & 0.243 (0.003) & 0.941 (0.003) & 0.950 (0.002) \\ \hline

\textbf{DRES + KNORA-E}    & 0.371 (0.003) & \textbf{0.973 (0.002)*} & 0.986 (0.002)* \\
\textbf{DRES + META-DES}   & 0.367 (0.002) & 0.972 (0.002)* & 0.980 (0.005) \\
\textbf{DRES + DES-P}      & \textbf{0.385 (0.003)*} & 0.972 (0.003)* & 0.984 (0.002)* \\
\hline
\end{tabular}
}
\label{tab:end2end_transformers}
\end{table}
As shown in Table~\ref{tab:end2end_transformers}, DRES consistently outperforms both end-to-end fine-tuned LLMs and static ensemble strategies across all datasets. While recent models such as LLaMA3 and Mistral achieve strong F1 scores, particularly on the GM and COVID datasets, the DRES variants surpass them by a clear margin, with gains of up to 11 F1 points on the LIAR dataset and around 2 to 4 points on the others. In contrast, static ensemble baselines (Groups A, B, and C) do not outperform the best fine-tuned LLMs, indicating that simple model aggregation provides limited benefits in this setting. These results emphasize the effectiveness of DRES's dynamic selection strategy in addressing instance-specific challenges.

It is important to acknowledge the possibility of training data contamination for recently released language models such as LLaMA3 and Mistral. These models were made available after the release of the datasets used in this study (LIAR, COVID, GM), suggesting that portions of these datasets may have been included in their pretraining corpora. This raises concerns about potential memorization effects, particularly in fine-tuned or zero-shot scenarios.

\begin{figure}[t]
    \begin{subfigure}[b]{1\linewidth}
        \centering
        \includegraphics[width=\linewidth]{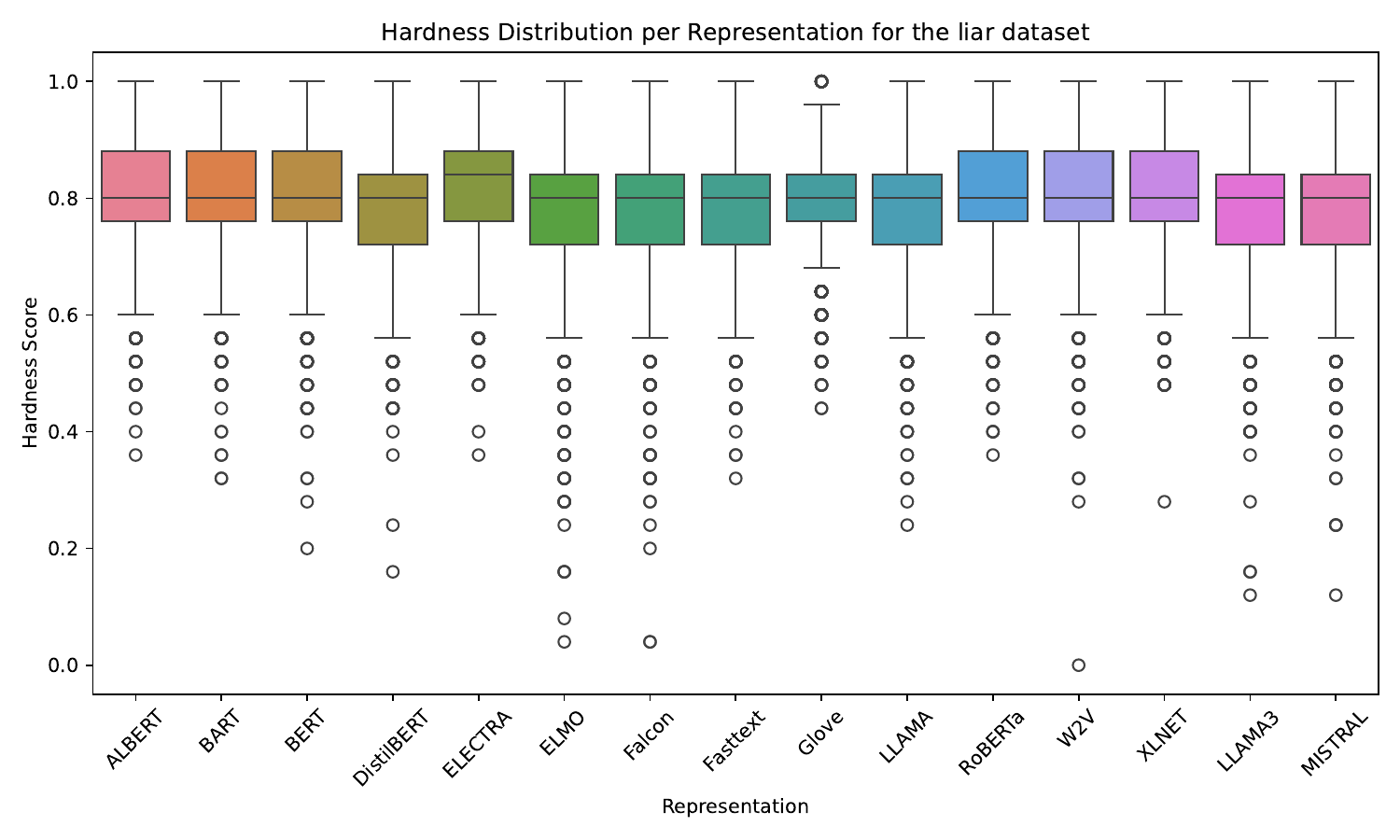}
        \caption{Liar Dataset}
        \label{fig:boxplotliar}
    \end{subfigure}

    \begin{subfigure}[b]{1\linewidth}
        \centering
        \includegraphics[width=\linewidth]{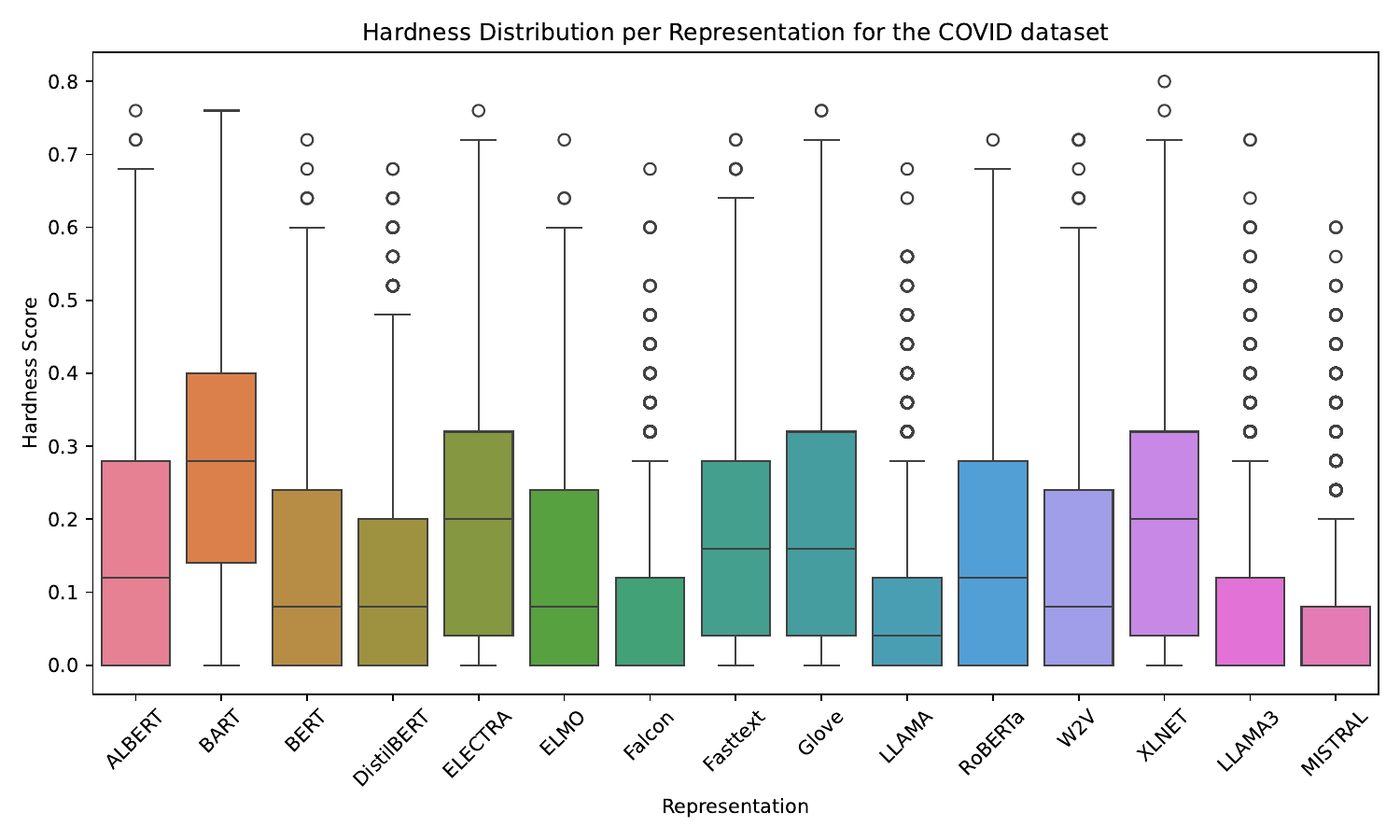}
        \caption{COVID Dataset}
        \label{fig:boxplotcovid}
    \end{subfigure}

    \begin{subfigure}[b]{1\linewidth}
        \centering
        \includegraphics[width=\linewidth]{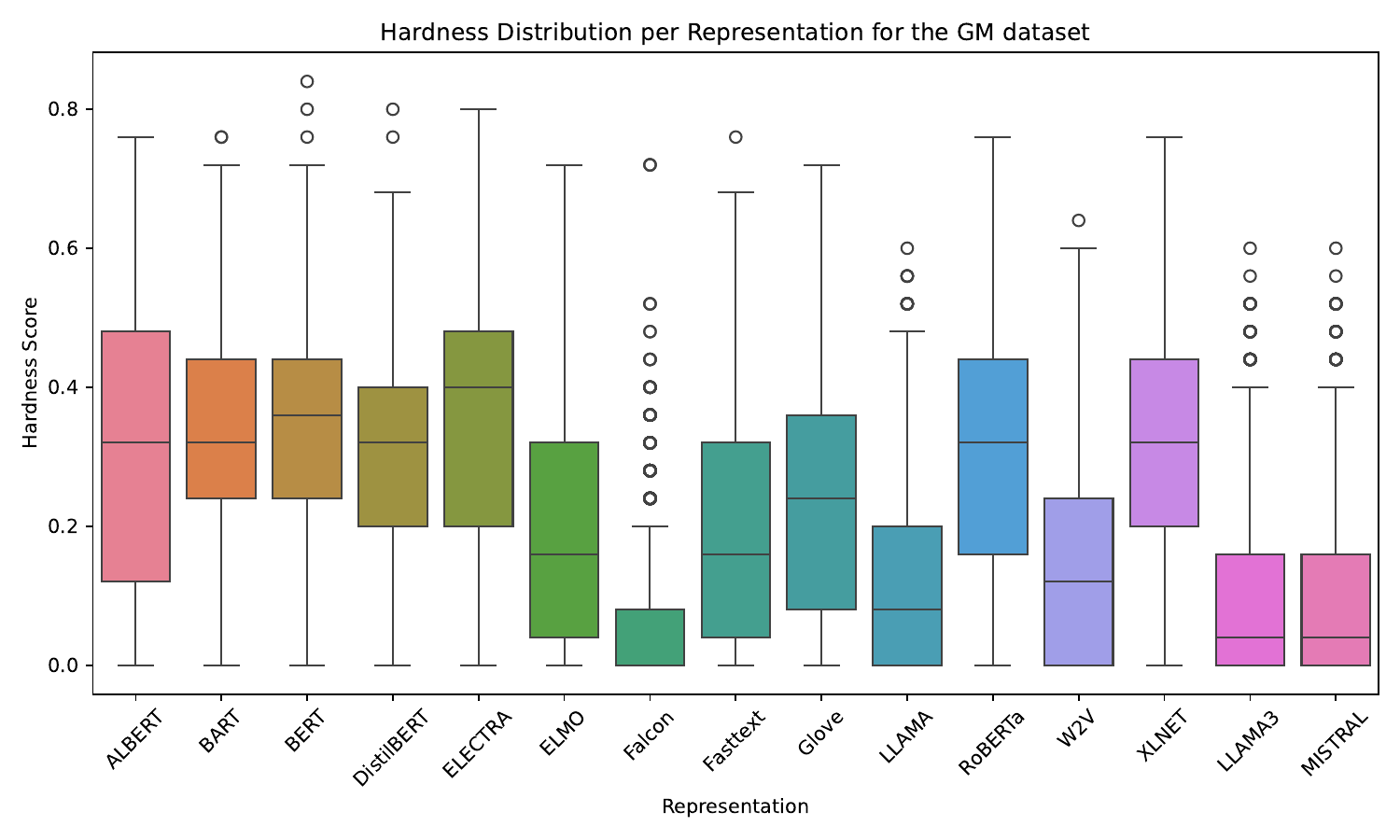}
        \caption{GM Dataset}
        \label{fig:boxplotgm}
    \end{subfigure}
        \caption{Boxplot showing the hardness distribution for the Liar, COVID and GM datasets.}
        \label{fig:boxplotdistribution}
\end{figure}

\section{Instance Hardness Distribution Analysis}
\label{Instance Hardness Distribution Analysis}

\begin{figure*}[htb]
    \centering
    \begin{subfigure}[b]{\textwidth}
        \centering
        \includegraphics[width=1\textwidth]{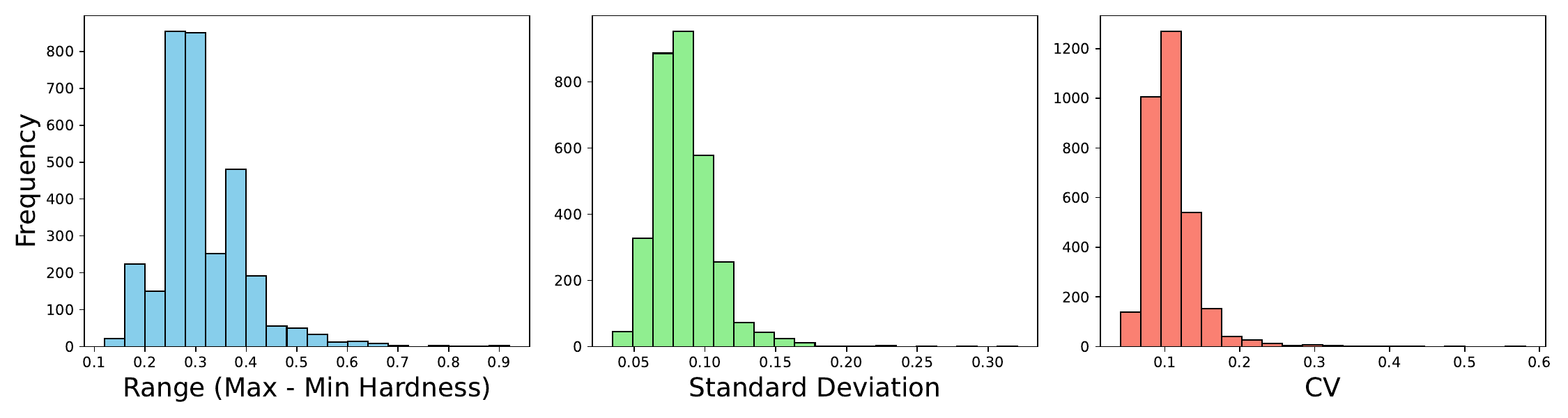}
        \caption{Liar dataset}
        \label{fig:stats1}
    \end{subfigure}
    
    \begin{subfigure}[b]{\textwidth}
        \centering
        \includegraphics[width=1\textwidth]{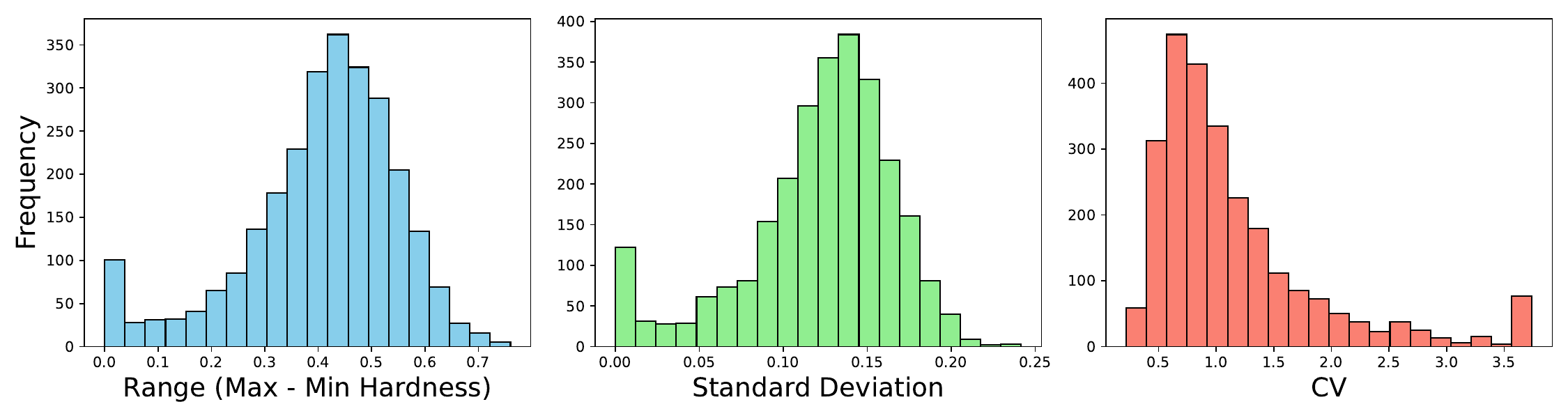}
        \caption{COVID dataset}
        \label{fig:stats2}
    \end{subfigure}
  
    \begin{subfigure}[b]{\textwidth}
        \includegraphics[width=\textwidth]{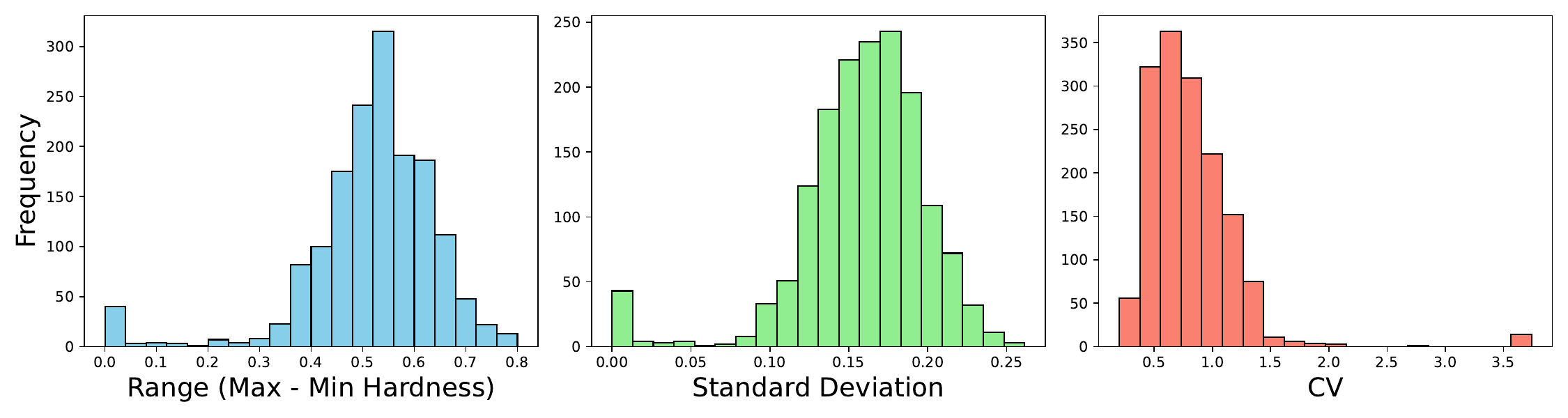}
        \caption{GM dataset}

        \label{fig:stats3}
    \end{subfigure}
    \caption{Descriptive statistics showing (left) range, (center) standard deviation, and (right) coefficient of variation for each dataset.}
    \label{fig:stats}
\end{figure*}

This section presents the distribution of instance hardness (IH) across various text representations and datasets. As shown in Figure~\ref{fig:boxplotdistribution}, IH scores vary not only between datasets but also across representations within the same dataset. These distributions reflect differences in data complexity. For instance, GM exhibits overall lower hardness values than LIAR, while COVID displays greater variability depending on the chosen feature space. Such variation helps explain why static combination strategies often fall short—some representations may work well globally, but others introduce noise or redundancy when applied uniformly.

These plots offer a global view of how separable or difficult samples are under each representation, supporting the idea that representation quality is highly data- and model-dependent. However, while useful for assessing overall trends, these distributions do not capture per-instance variation. That is, a representation with good average performance might still perform poorly on specific inputs. These analyses are conducted in the following section.

\section{Analysis of Hardness Variation Across Representations}
\label{app:variationrepresentation}

We analyze how instance hardness varies across different text representations using two complementary perspectives. First, we compute range, variance, and coefficient of variation (CV) as summary statistics to capture the dispersion of hardness values across representations for each instance. Then, we present a sorted hardness gap profile to visualize the differences in maximum and minimum hardness values across views.

\begin{figure}[h]
    \centering
    \begin{subfigure}[b]{1\linewidth}
        \centering
        \includegraphics[width=\linewidth]{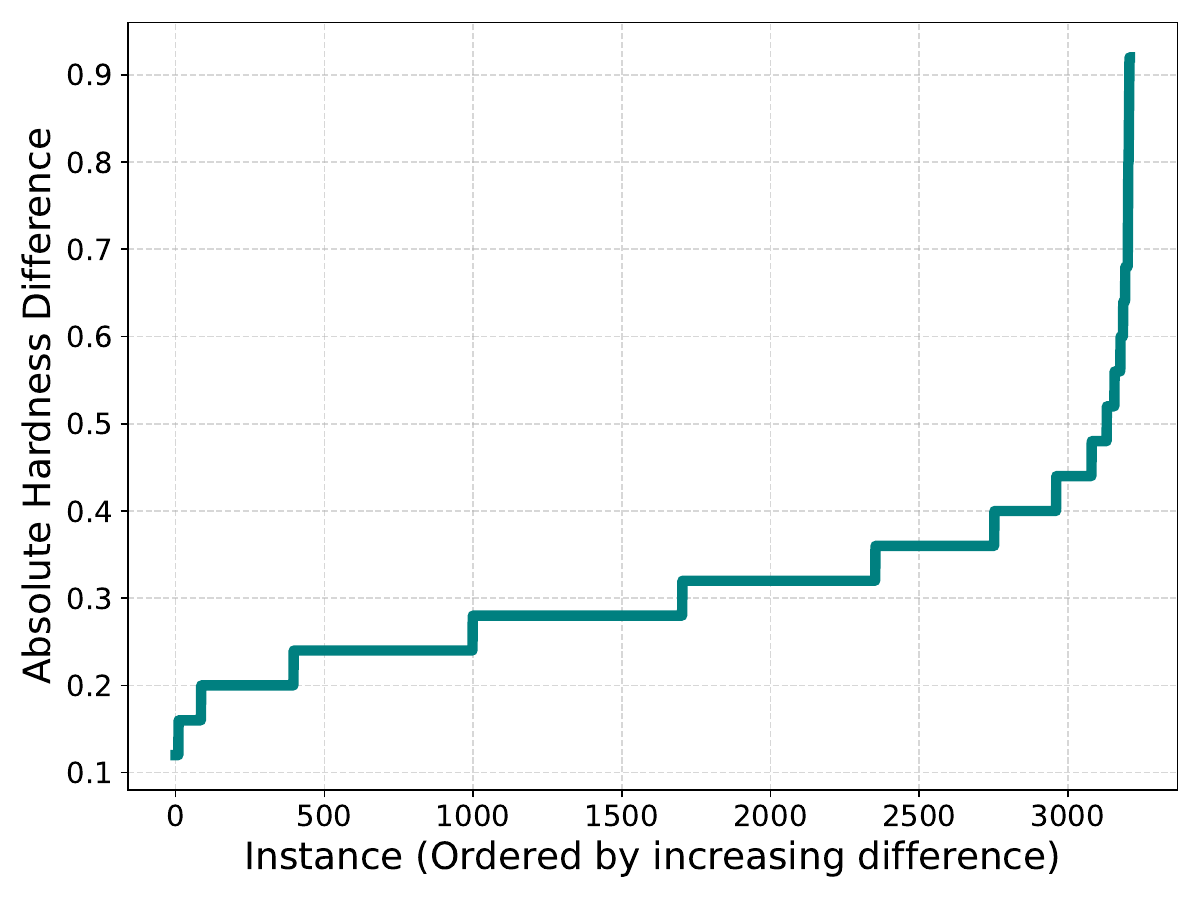}
        \caption{Liar Dataset}
    \end{subfigure}
        \begin{subfigure}[b]{1\linewidth}
        \centering
        \includegraphics[width=\linewidth]{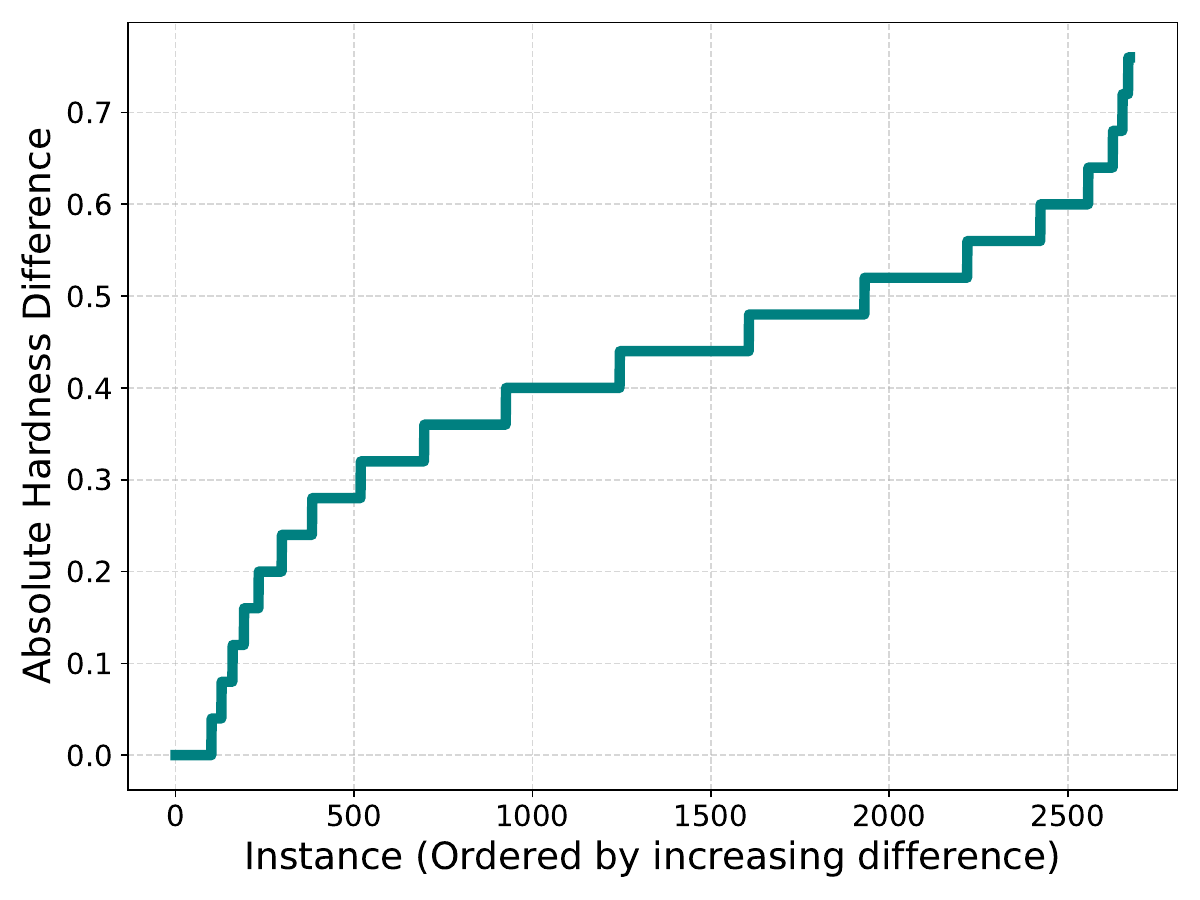}
        \caption{COVID Dataset}
    \end{subfigure}

     \begin{subfigure}[b]{1\linewidth}
        \centering
        \includegraphics[width=\linewidth]{New_Figures/ascending_difference_GM.pdf}
        \caption{GM Dataset}
    \end{subfigure}
        \caption{Cumulative distribution of the difference between maximum and minimum IH values across datasets. \label{fig:ascending}}
\end{figure}

As shown in Figure~\ref{fig:stats}, the range is defined as the difference between the maximum and minimum hardness scores per instance. The \textsc{Liar} dataset exhibits a broader spread, with most values under 0.5 but a long tail approaching 0.9. The \textsc{GM} and \textsc{COVID} datasets show tighter distributions, with most instances higher than 0.4, indicating that there is a significant hardness difference across representations for the majority of cases. 

In addition to summary statistics, Figure~\ref{fig:ascending} shows instance-wise hardness profiles sorted in ascending order of the gap between the maximum and minimum hardness scores. These plots reveal substantial disagreement across views. For instance, in the \textsc{GM} dataset, over 50\% of instances show a gap of at least 0.5, and 25\% exceed 0.7. The \textsc{COVID} and \textsc{Liar} datasets follow similar patterns, with significant fractions of instances showing gaps larger than 0.4.

These findings further support the motivation behind our dynamic instance hardness representation selection approach. Instead of relying on a fixed or fused representation, DRES uses test-time hardness estimation to adaptively select the most appropriate view. The observed variation in hardness supports this strategy, as many instances would be suboptimally handled by any single representation or by combining all representations.

\section{Instance Hardness Heatmaps}
\label{app:hardnessheatmap}

Figure~\ref{fig:heatmaps} shows heatmaps of instance hardness scores across all representations for the LIAR, COVID, and GM datasets. Each row corresponds to a representation, and each column to a sample. The LIAR dataset displays consistently higher hardness values across representations, with many samples appearing difficult to classify regardless of the embedding space. In contrast, COVID and GM show larger regions of low hardness, indicating more apparent class separation and more consistent behavior across representations. These patterns help explain the results in Table~\ref{tab:dmes_ablation}, where LIAR shows the most significant performance gain (+5.6 percentage points) when using dynamic ensemble selection (DES) alone. This improvement can be explained by the DES’s ability to deal with high disagreement between classifiers and better handle harder instances~\cite{cruz2017dynamic}.

\begin{figure*}[htb]
    \centering
    \begin{subfigure}[b]{1\linewidth}
        \centering
        \includegraphics[width=\linewidth]{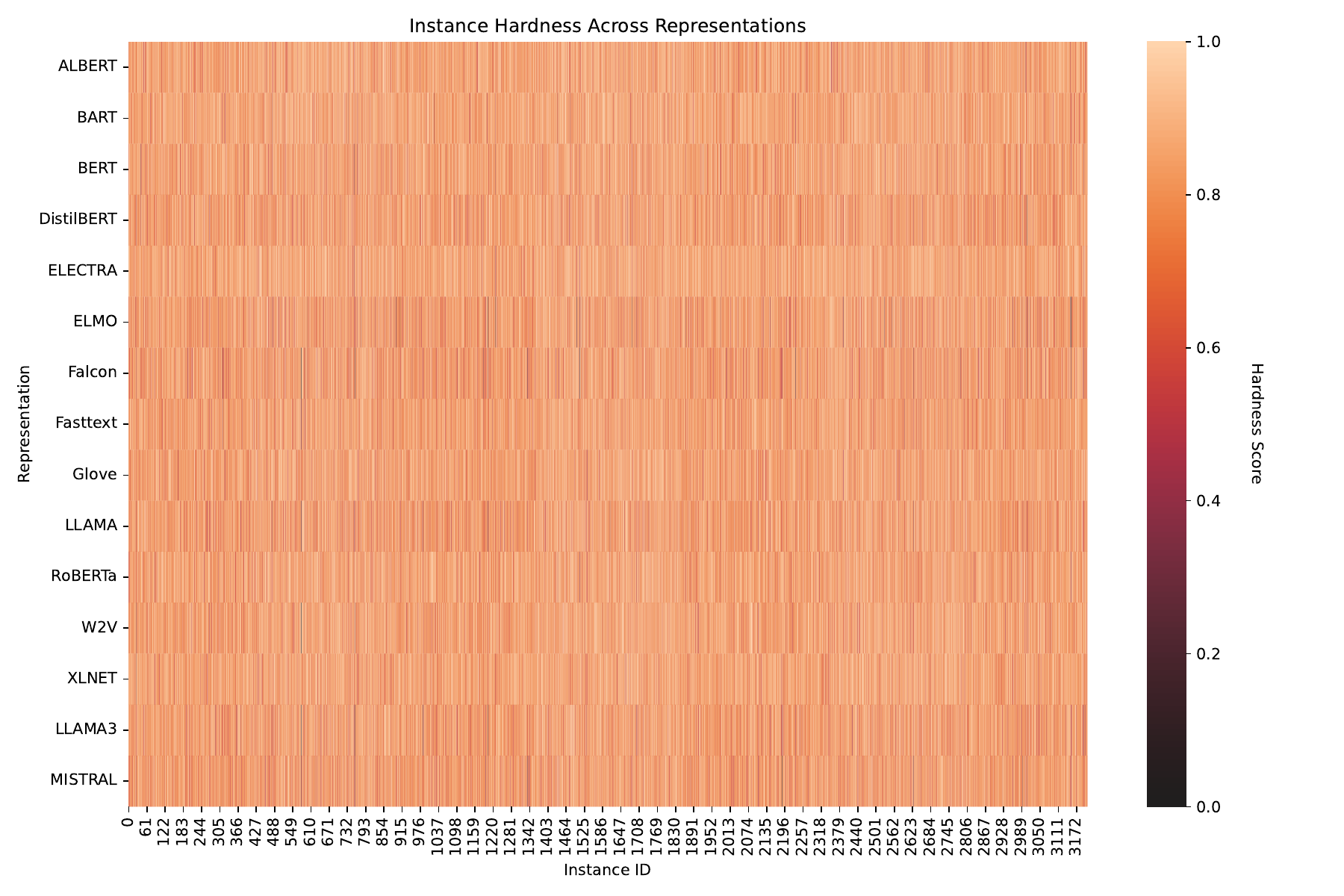}
        \caption{Liar dataset}
        \label{fig:heat1}
    \end{subfigure}
    \hfill
    \begin{subfigure}[b]{1\linewidth}
        \centering
        \includegraphics[width=\linewidth]{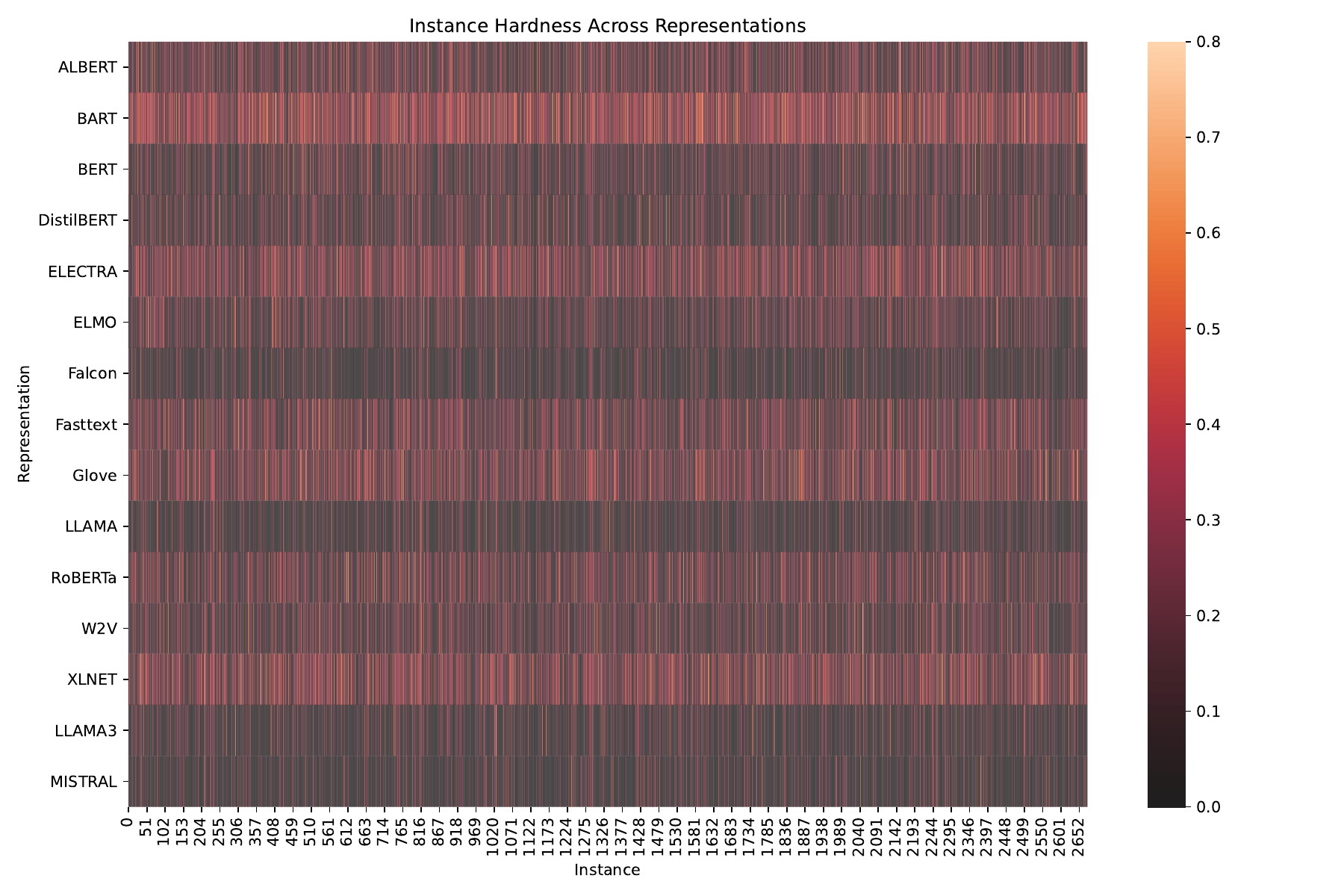}
        \caption{COVID dataset}
        \label{fig:heat2}
    \end{subfigure}
    \caption{Instance hardness heatmaps across datasets (a)–(b).}
    \label{fig:heatmaps}
\end{figure*}

\begin{figure*}[ht]
    \ContinuedFloat
    \centering
    \begin{subfigure}[b]{1\linewidth}
        \centering
        \includegraphics[width=\linewidth]{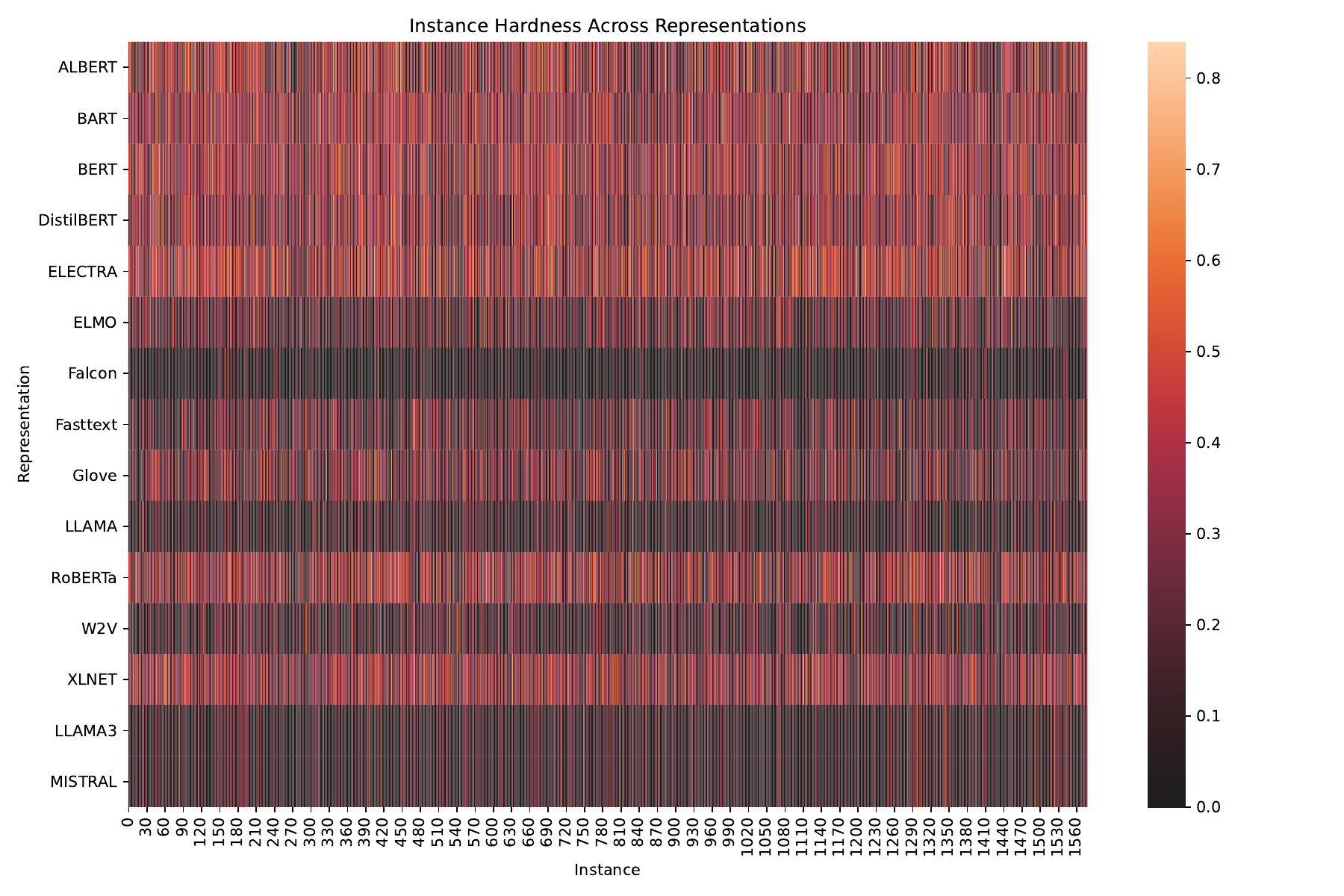}
        \caption{GM dataset}
        \label{fig:heat3}
    \end{subfigure}
    \caption{(Continued) Instance hardness heatmaps across datasets (c).}
\end{figure*}

\section{Usage of AI Assistants}
The authors used AI tools (ChatGPT and Grammarly) to help the manuscript writing process, specifically for revising grammar, improving clarity, and checking mathematical notation consistency. AI was also used to assist with coding tasks for data analysis and visualizations (e.g., heatmaps, radar plots, and cumulative distributions). All content and code were reviewed, verified, and finalized by the authors.

\end{document}